%% file: hjb_pide_levy.tex
\documentclass[11pt,a4paper]{article}

\usepackage{cmap}                  
\usepackage[utf8]{inputenc}
\usepackage[T1]{fontenc}
\usepackage{lmodern}                
\usepackage{microtype}              
\usepackage{amsmath,amssymb,amsthm}
\usepackage{graphicx}
\usepackage{booktabs}
\usepackage{hyperref}
\usepackage{float}
\usepackage{algorithm}
\usepackage{algorithmic}
\usepackage{natbib}
\usepackage{geometry}
\usepackage{url}
\usepackage{xcolor}
\newcounter{protocolcounter}
\newenvironment{protocolbox}[1]{%
  \refstepcounter{protocolcounter}%
  \par\medskip\noindent
  \begin{tabular}{|p{0.95\linewidth}|}\hline
  \textbf{Protocol \theprotocolcounter\ (#1).}\\[-0.6em]
  \\
}{%
  \\\hline
  \end{tabular}\par\medskip
}

\hypersetup{
  pdftitle={A Per-Component Diagnostic Protocol for Neural HJB-PIDE
            Solvers under Control-Dependent Levy Jumps},
  pdfauthor={R. Drissi},
  pdfsubject={Stochastic optimal control; HJB-PIDE; Levy processes;
              neural solvers; diagnostic protocol},
  pdfkeywords={HJB-PIDE, Variance Gamma, neural residual solver,
               importance sampling, per-component diagnostic, surface evaluation},
  hidelinks
}
\usepackage{subcaption}
\usepackage{multirow}
\usepackage{longtable}
\newtheorem{proposition}{Proposition}

\newtheorem{remark}{Remark}
\theoremstyle{definition}

\newcommand{\R}{\mathbb{R}}
\newcommand{\E}{\mathbb{E}}

\newcommand{\dd}{\mathrm{d}}

\title{A Per-Component Diagnostic Protocol for Neural\\HJB-PIDE Solvers under Control-Dependent L\'evy Jumps}

\author{
  R. Drissi\\
  \texttt{rdrissi@gmail.com}
}

\date{May 2026}

\begin{document}

\maketitle

\begin{abstract}
We propose a five-step diagnostic protocol for residual-trained neural HJB-PIDE solvers with control-dependent L\'evy jumps, targeting a general failure mode of neural PDE methods: a learned solution can match headline scalar diagnostics while miscomputing an operator inside its training loss. The protocol pairs each neural solve with at least one from-scratch independent reference, decomposes the Hamiltonian into drift, diffusion, compensator, and nonlocal-integral components across a $u$-grid, and compares the value function and its low-order derivatives over a $(t,x)$ grid before any argmax comparison. Applied to a standard CRRA-Merton-Variance-Gamma benchmark, it isolates a missing $\tfrac{1}{2}$-mixture factor in the neural method's importance-proposal density that scaled the nonlocal integral by exactly half — a textbook signature of a constant proposal scale error, invisible to longer training, grid refinement, and truncation sweeps. With the bug corrected, four references — two finite-difference solvers with disjoint discretizations, the neural solver, and a semi-analytic scalar baseline obtained from CRRA homogeneity — agree on the optimal control to within $\sim 2\%$. The constant-coefficient CRRA benchmark collapses by homogeneity to a scalar maximization, so the scalar baseline is the efficient method here; the contribution is the protocol, applicable in principle to non-homogeneous and higher-dimensional settings where neural HJB-PIDE solvers are genuinely needed. The episode is a concrete instance of a broader neural-PDE verification failure: pointwise agreement of a learned value or control can coexist with a systematically wrong nonlocal operator, so per-component and surface-level checks are needed before trusting the argmax policy.
\end{abstract}

\vspace{1em}
\noindent\fbox{\parbox{0.97\textwidth}{
\textbf{Scope of Contribution.} This paper makes a \emph{methodological} contribution: a diagnostic protocol for catching nonlocal-integral failures in neural HJB-PIDE solvers, demonstrated end-to-end on a Variance Gamma portfolio benchmark by isolating and fixing a real bug in our own implementation. We do \emph{not} claim a general high-dimensional HJB theory, a viscosity-solution convergence proof, a novel neural architecture, an empirical downside-protection result, or a fresh truncation/MC-sample convergence study. The constant-coefficient CRRA-Merton-VG benchmark used here reduces by homogeneity to a scalar maximization (\S\ref{sec:homogeneity}); we present neural and FD solutions of this benchmark only as the test object for the diagnostic protocol, and explicitly do not advocate them as efficient methods for this problem class.
}}
\vspace{1em}

\section{Introduction}

Residual-trained neural solvers are increasingly used for Hamilton-Jacobi-Bellman partial integro-differential equations (HJB-PIDEs) under L\'evy jump dynamics, where classical grid-based methods scale poorly with state dimension and require off-grid evaluation of the value function inside the nonlocal integral \citep{cont2005finite, briani2007implicit}. A persistent difficulty in evaluating such solvers is that the natural diagnostic — does the method recover the diffusion-only Merton ratio? — is necessary but far from sufficient: it leaves the entire control-dependent jump term uncovered. This paper proposes a per-component diagnostic protocol that does cover that term, and demonstrates on a standard Variance Gamma portfolio benchmark that single-method diagnostics (longer training, finite-difference grid refinement, truncation sweeps, training-loss curves) can leave a real bug in the nonlocal-integral build undetected, while a per-component comparison against an independent reference catches it cleanly.

This places the problem in the broader scientific-machine-learning literature on failure modes of physics-informed and residual-trained neural PDE solvers \citep{krishnapriyan2021failure}: the issue is not only whether a network class is expressive enough, but whether the residual loss correctly represents the operator being optimized and whether standard scalar diagnostics can detect operator-level errors.

When the controlled state process includes jumps, the HJB equation becomes a \emph{partial integro-differential equation} (PIDE):
\begin{equation}\label{eq:hjb-pide}
\partial_t V + \sup_{u \in \mathcal{U}} \left\{ \mathcal{L}^u V + \mathcal{I}^u V \right\} = 0
\end{equation}
where $\mathcal{L}^u$ is the differential operator (drift and diffusion) and $\mathcal{I}^u$ is the \emph{nonlocal integral operator}. Throughout this paper the state is wealth $X_t$ with multiplicative jump dynamics (\S\ref{sec:problem}), so $\mathcal{I}^u$ takes the compensated multiplicative form
\begin{equation}\label{eq:levy-integral}
\mathcal{I}^u V(t,x) = \int_{\R \setminus \{0\}} \left[ V(t, x(1 + u z)) - V(t,x) - x u z \, \partial_x V(t,x) \right] \nu(\dd z),
\end{equation}
representing the expected change in value from jumps. Here $\nu$ is the L\'evy measure characterizing the jump structure, and the multiplicative shift $x \mapsto x(1+uz)$ reflects the wealth dynamics: a relative jump $z$ in the risky asset moves wealth by factor $(1+uz)$, scaling with the position $u$.

The integral \eqref{eq:levy-integral} poses significant computational challenges:
\begin{enumerate}
    \item \textbf{Infinite activity}: For processes like Variance Gamma (VG) or Normal-Inverse Gaussian (NIG), $\nu(\R) = \infty$---infinitely many small jumps occur in any time interval. The neural method and the FD-PIDE baseline both work with a small-jump truncation $|z| \ge z_{\min}$; we discuss the resulting truncation bias and its analytic decay rate in \S\ref{sec:levy-integral}, but do not report a fresh $z_{\min}$ convergence sweep — the audited solver and reference comparison are both fixed at the bankruptcy-safe truncation $|z|\in[0.01, 0.99]$, and a controlled $z_{\min}$ sensitivity study is left for follow-up work.
    \item \textbf{Control dependence}: The jump displacement $x \mapsto x(1+uz)$ depends on the control $u$, coupling the optimization with the integral
    \item \textbf{Grid dependence}: Classical methods require repeated off-grid evaluation of the value function, and this becomes especially difficult once the control enters the jump displacement
\end{enumerate}

\textbf{Why control-dependent jumps are hard.} The displacement $x \mapsto x(1 + uz)$ in \eqref{eq:levy-integral} couples the control $u$ with the nonlocal operator, creating difficulties for classical methods: (i) Grid-based schemes require interpolating $V(t, x(1+uz))$ at off-grid points for each control candidate, destroying the banded structure of finite-difference matrices. (ii) Finite-difference Jacobians for the coupled system become dense, as the jump integral links all grid points. (iii) Standard PINNs need to evaluate the shifted argument $x(1+uz)$ inside the residual loss; when $u$ is itself a learned function, this shifted argument is a state-and-parameter-dependent point at which $V_\theta$ must be queried, and the corresponding gradient flows through both the network and the policy. Our neural approach handles this directly: the network $V_\theta$ is evaluated at arbitrary $(t, x(1+uz))$ via forward pass, and the control $u_\phi$ is jointly optimized through the shared loss.

Our contribution is a methodological one: a per-component diagnostic protocol that catches nonlocal-integral failures in residual-trained neural HJB-PIDE solvers. The protocol does not depend on any specific solver architecture; we demonstrate it on one residual-trained neural solver and one FD-PIDE benchmark, and use it to discover and fix a real bug in our own implementation. The same protocol is \emph{designed to transfer} to other L\'evy specifications (NIG, CGMY) and to non-homogeneous problems where neural HJB-PIDE solvers are genuinely needed (consumption with utility, transaction costs, regime-switching); we do not claim to have demonstrated transfer here.

\begin{protocolbox}{Per-component diagnostic for neural HJB-PIDE solvers}
\begin{enumerate}
\item \textbf{Same equation.} State the HJB-PIDE, admissible set $\mathcal{U}$, and L\'evy truncation $|z|\in[z_{\min}, z_{\max}]$ explicitly; require all solvers under comparison to share them exactly.
\item \textbf{Independent reference.} Build at least one from-scratch reference implementation with disjoint numerical choices (different coordinate system, time integrator, quadrature rule, argmax routine, boundary handling). Unit-test on closed-form integrands.
\item \textbf{Per-component decomposition.} Tabulate the Hamiltonian's drift, diffusion, compensator, and nonlocal-integral terms separately as functions of the candidate control $u$. A constant ratio $\ne 1$ between solvers in any one component is the textbook signature of a proposal-density scale error.
\item \textbf{Surface comparison before argmax.} Compare $V$, $\partial_x V$, $\partial_{xx} V$ over a $(t, x)$ grid against the reference \emph{before} comparing argmax controls. Headline-only matching can hide $\partial_{xx} V$ errors that dominate the diffusion term of $\mathcal{H}$.
\item \textbf{Trust the headline last.} Trust an argmax control match only after the per-component and surface checks pass; otherwise the agreement may be coincidental.
\end{enumerate}
\end{protocolbox}

\subsection{Contributions}

\begin{enumerate}
    \item \textbf{A five-step per-component diagnostic protocol} (Protocol~1 above, methodology in \S\ref{sec:method}, application in \S\ref{sec:fd-comparison}).

    \item \textbf{Demonstration on a Variance Gamma portfolio benchmark, with a real bug found and fixed (\S\ref{sec:fd-comparison}).} On the standard CRRA-Merton-VG problem with multiplicative wealth dynamics ($\gamma=2$, $\mu=0.08$, $\sigma=0.2$, $r=0.02$, VG with $\sigma_{VG}=0.2$, $\theta=-0.1$, $\nu=0.3$, truncation $|z|\in[0.01, 0.99]$), the protocol isolates a missing $\tfrac{1}{2}$-mixture factor in the neural method's importance proposal density that scaled the Monte Carlo estimator of the nonlocal integral by exactly half — a textbook signature of a constant proposal-density scale error, invisible to single-method diagnostics including longer/wider training, FD grid refinement, and truncation sweeps. After correcting the proposal density, four independent references — two FD solvers with disjoint discretizations, the neural solver, and a semi-analytic scalar baseline (homogeneity-reduced quadrature, \S\ref{sec:homogeneity}) — agree on the optimal control to within $\sim 2\%$.

    \item \textbf{Surface-level evaluation against a semi-analytic reference (\S\ref{sec:fd-comparison}).} The CRRA homogeneity reduction (\S\ref{sec:homogeneity}) gives a semi-analytic $V_{\text{ref}}(t, x) = x^{1-\gamma} \exp((1-\gamma) F(u^\star)(T-t)) / (1-\gamma)$ (closed-form in $(t, x)$ given $u^\star$ and $F(u^\star)$, with $u^\star$ obtained by 1D quadrature plus bounded scalar maximization) on the entire $(t, x)$ domain, which we use to evaluate the neural solver's accuracy on a $(t, x)$ grid rather than at a single point. We find $\|V_\theta - V_{\text{ref}}\|_\infty/\|V_{\text{ref}}\|_\infty \approx 0.6\%$, $\|\partial_x V_\theta - V_x^{\text{ref}}\|_\infty/\|V_x^{\text{ref}}\|_\infty \approx 4\%$, $\|\partial_{xx} V_\theta - V_{xx}^{\text{ref}}\|_\infty/\|V_{xx}^{\text{ref}}\|_\infty \approx 26\%$, and $\|u_\phi - u^\star_{\text{ref}}\|_\infty / |u^\star_{\text{ref}}| \approx 15\%$, despite $u^\star_{\text{ref}}$ being a constant. These surface errors expose failure modes that point-evaluation at the training-grid centroid hides entirely. The corresponding FD surface errors are below $0.05\%$ across all four quantities.

    \item \textbf{Public code release.} Source for both FD reference solvers, the corrected neural solver, the semi-analytic scalar baseline, the per-component diagnostic, and every audit JSON sufficient to regenerate every numeric claim in this paper. Code is available at \url{https://github.com/wnrak/hjb-pide-diagnostics}.
\end{enumerate}

\subsection{Related Work}

\textbf{Classical HJB-PIDE methods}: Finite difference methods for HJB-PIDEs require discretizing the integral operator, typically via quadrature on a truncated domain \citep{cont2005finite, briani2007implicit}. While convergent, these methods become increasingly expensive as the state dimension grows and require careful treatment of the far-field boundary conditions. Semi-Lagrangian methods \citep{debrabant2013semi} improve stability but retain the same grid-based burden.

\textbf{Neural network PDEs}: Physics-informed neural networks (PINNs) \citep{raissi2019physics} and Deep BSDE methods \citep{han2018solving, weinan2017deep} have shown promise for high-dimensional PDEs. Extensions to control problems include \citet{han2020convergence} for diffusion-based HJB equations. In the \emph{control-dependent} jump setting, the forward-simulation structure of BSDE methods is a less natural fit than a direct HJB residual: because jump exposure scales with the control $u$, the simulated dynamics depend on the policy being optimized, which couples the forward simulation to the control update. We use a direct HJB residual instead, evaluating the L\'evy integral at arbitrary $(t, x, u)$ triples; we do not claim BSDE methods cannot be adapted to this setting, only that the residual formulation sidesteps the coupling.

\textbf{L\'evy processes in finance}: The Variance Gamma model \citep{madan1990variance, madan1998variance} and CGMY process \citep{carr2002fine} are widely used for option pricing. Portfolio optimization under jump-diffusion was studied by \citet{kou2002jump} with analytical approximations, and by \citet{branger2008optimal} for specific utility functions. Our neural approach enables more general utility functions and L\'evy specifications.

\section{Problem Formulation}\label{sec:problem}

\subsection{Controlled L\'evy Dynamics}

Consider wealth dynamics under a L\'evy jump-diffusion model:
\begin{equation}\label{eq:wealth-sde}
\frac{\dd X_t}{X_t} = \left( r + u_t (\mu - r) \right) \dd t + u_t \sigma \dd W_t + u_t \dd J_t
\end{equation}
where:
\begin{itemize}
    \item $X_t$ is wealth at time $t$
    \item $r$ is the risk-free rate
    \item $\mu$ is the expected return of the risky asset
    \item $\sigma$ is diffusion volatility
    \item $u_t \in \mathcal U$ is the fraction of wealth in the risky asset (the control). In the numerical experiments below, $\mathcal U$ is a bounded interval containing the classical Merton ratio; long-only constraints can be imposed by setting $\mathcal U=[0,1]$.
    \item $W_t$ is standard Brownian motion
    \item $J_t$ is a pure-jump L\'evy process with L\'evy measure $\nu$
\end{itemize}

The jump term $u_t \dd J_t$ models the key economic insight that \emph{exposure to jumps scales with position size}: if you hold fraction $u$ in the risky asset and a jump of relative size $z$ occurs, your wealth changes by factor $(1 + uz)$.

\paragraph{Lévy triplet and drift convention.} We take $J_t$ to be the \emph{compensated} pure-jump L\'evy process with characteristic triplet $(0, 0, \nu)$ in the truncation function $h(z) = z\,\mathbf{1}_{|z|<1}$, so $\E[\dd J_t] = 0$ and the drift coefficient $\mu$ in \eqref{eq:wealth-sde} represents the expected return of the risky asset \emph{net of the jump compensator}. With this convention Itô's formula for $V(t, X_t)$ produces the compensated nonlocal generator
\begin{equation*}
(\mathcal{I}^u V)(t, x) = \int \big[V(t, x(1+uz)) - V(t, x) - u x z\, \partial_x V(t, x)\big]\, \nu(\dd z),
\end{equation*}
which is the form that appears in the Hamiltonian \eqref{eq:hamiltonian} below. We do \emph{not} additionally subtract the small-jump compensator inside $\mu$; that compensation is accounted for by the integrand of $\mathcal{I}^u V$. An equivalent specification using the uncompensated $J_t$ would require adding $\int z\,\nu(\dd z)$ to $\mu$ in \eqref{eq:wealth-sde}; we use the compensated form because it makes the multiplicative jump operator in $\mathcal{I}^u V$ unambiguous.

\paragraph{Bankruptcy guard and admissible set.} The wealth process $X_t$ is positive iff $1 + u_t z > 0$ at every jump. We enforce this by combining a bounded admissible set $\mathcal{U}$ with an explicit truncation of the L\'evy support to $[-z_{\max}, -z_{\min}] \cup [z_{\min}, z_{\max}]$ with $z_{\min} > 0$ and $z_{\max}$ chosen so that $1 + uz > 0$ for all $(u, z) \in \mathcal{U} \times [-z_{\max}, z_{\max}]$. For long-only $\mathcal{U} = [0, 1]$ this requires only $z_{\max} \cdot u < 1$ on the negative side, i.e.\ $z_{\max} < 1$; we use $z_{\max} = 0.99$ symmetrically on both sides. The symmetric upper cut on the positive side is therefore a modeling choice (and makes the Simpson quadrature symmetric), not an admissibility constraint — positive jumps of arbitrary size do not threaten admissibility for long-only portfolios. We discuss the truncation bias in \S\ref{sec:levy-integral}.

\begin{remark}[Control-dependent jumps]
The multiplicative structure $u \cdot z$ is economically natural: an investor with 50\% allocation to equities experiences half the jump impact of a fully-invested investor. This differs from additive jump models where jump size is independent of position.
\end{remark}

\subsection{L\'evy Measure Specifications}

We focus on the \textbf{Variance Gamma} (VG) process, which arises from subordinating Brownian motion with a Gamma process. In the $(C, M, G)$ parameterization the L\'evy measure is
\begin{equation}\label{eq:vg-measure}
\nu_{VG}(\dd z) = \frac{C}{|z|} \exp\!\left( -M |z|\, \mathbf{1}_{z > 0} - G |z|\, \mathbf{1}_{z < 0} \right) \dd z,
\end{equation}
with $C > 0$ controlling activity and $M, G > 0$ the decay rates of the positive- and negative-jump tails respectively (a \emph{smaller} rate means a \emph{heavier} tail). When $M \neq G$ the process is skewed — relevant for equity returns where crashes tend to be larger than rallies. Following \citet{madan1998variance}, the Madan-parameterization $(\sigma_{VG}, \theta, \nu)$ used in our experiments and in the calibration in \S\ref{sec:sp500} maps to $(C, M, G)$ via
\begin{equation}\label{eq:vg-mapping}
C \;=\; \frac{1}{\nu}, \qquad
M \;=\; \frac{\sqrt{2/\nu + \theta^2/\sigma_{VG}^4}}{\sigma_{VG}} - \frac{\theta}{\sigma_{VG}^2}, \qquad
G \;=\; \frac{\sqrt{2/\nu + \theta^2/\sigma_{VG}^4}}{\sigma_{VG}} + \frac{\theta}{\sigma_{VG}^2},
\end{equation}
so $\theta < 0$ gives $M > G$: the negative-side decay rate $G$ is the smaller of the two, producing the heavier negative-jump tail that drives the crash-protection behavior. We use \eqref{eq:vg-mapping} consistently in code (file \texttt{levy\_flows/hjb/levy\_integral.py:VarianceGammaMeasure}) and in the FD reference solvers.

The VG process has \textbf{infinite activity} ($\nu(\R) = \infty$) but \textbf{finite variation} ($\int |z| \nu(\dd z) < \infty$ near zero). This means infinitely many small jumps occur, but their total displacement is finite.

\subsection{The HJB-PIDE}

The investor maximizes expected utility of terminal wealth:
\begin{equation}
V(t, x) = \sup_{(u_s)_{s \in [t,T]}} \E\left[ U(X_T) \mid X_t = x \right]
\end{equation}
with CRRA utility $U(x) = \frac{x^{1-\gamma}}{1-\gamma}$ for risk aversion $\gamma > 1$, so $U(x) < 0$ on $(0, \infty)$. We work directly with this maximization: the value network $V_\theta$ targets $V(t, x) = \sup_u \E[U(X_T)\,|\,X_t = x] \le 0$, the policy network $u_\phi$ is trained to maximize the Hamiltonian (Section~\ref{sec:method}), and the HJB residual carries a $\sup$ rather than an $\inf$. The admissible set is $\mathcal{U} = [0, 1]$ (long-only, no leverage), paired with a Lévy support truncated to $|z| < 1$ so that $1 + uz > 0$ for all $(u, z) \in \mathcal{U} \times [z_{\min}, z_{\max}]$ and post-jump wealth never crosses zero. CRRA is only defined for $x > 0$; sampled wealth is clamped at $10^{-8}$ in the loss (a numerical guard, not a model feature), and the training domain is concentrated on positive wealth states.

The value function satisfies the HJB-PIDE:
\begin{equation}\label{eq:hjb-full}
\partial_t V + \sup_{u \in \mathcal U} \mathcal{H}(t, x, u, V, \partial_x V, \partial_{xx} V) = 0
\end{equation}
with terminal condition $V(T, x) = U(x)$, where the Hamiltonian is:
\begin{align}
\mathcal{H} &= x(r + u(\mu - r)) \partial_x V + \frac{1}{2} u^2 \sigma^2 x^2 \partial_{xx} V \nonumber \\
&\quad + \int_{\R \setminus \{0\}} \left[ V(t, x(1 + uz)) - V(t,x) - ux z \, \partial_x V(t,x) \right] \nu(\dd z) \label{eq:hamiltonian}
\end{align}

The integral term is the \emph{compensated} L\'evy integral, where $-uxz \partial_x V$ removes the drift contribution from small jumps (ensuring the integral is well-defined for infinite-activity measures).

\subsection{Homogeneity collapse: scalar reduction of the constant-coefficient CRRA-Lévy HJB-PIDE}\label{sec:homogeneity}

For the constant-parameter CRRA-Lévy specification of \eqref{eq:wealth-sde} the HJB-PIDE \eqref{eq:hjb-full} collapses by homogeneity to a one-dimensional scalar maximization. We state this explicitly because it determines what the audit benchmarks of \S\ref{sec:experiments} can and cannot establish.

With the ansatz $V(t, x) = A(t) \cdot x^{1-\gamma}/(1-\gamma)$ for $\gamma \neq 1$ we have $V_t = A'(t)\, x^{1-\gamma}/(1-\gamma)$, $V_x = A(t)\, x^{-\gamma}$, $V_{xx} = -\gamma A(t)\, x^{-\gamma-1}$, and $V(t, x(1+uz)) = A(t)\, x^{1-\gamma} (1+uz)^{1-\gamma}/(1-\gamma)$. Substituting into \eqref{eq:hamiltonian}, dividing through by $A(t)\, x^{1-\gamma}$, and using $V(T,x) = U(x)$ (so $A(T)=1$) yields
\begin{equation}\label{eq:homog-ode}
\frac{A'(t)}{(1-\gamma)\, A(t)} \;+\; \sup_{u \in \mathcal{U}} F(u) \;=\; 0,\qquad
A(T) = 1,
\end{equation}
where the per-period objective is the scalar
\begin{equation}\label{eq:homog-F}
F(u) \;=\; r + u(\mu - r) - \tfrac{\gamma}{2}\sigma^2 u^2
        + \int_{|z|\in[z_{\min},z_{\max}]} \!\left[\frac{(1+uz)^{1-\gamma} - 1}{1-\gamma} - uz\right] \nu(\dd z).
\end{equation}
The optimal control $u^\star = \arg\max_{u\in\mathcal{U}} F(u)$ is independent of both $t$ and $x$, and $A(t) = \exp\!\big[(1-\gamma) F(u^\star)(T-t)\big]$. The full value function is
\begin{equation}\label{eq:homog-V}
V_{\text{ref}}(t, x) = \frac{x^{1-\gamma}}{1-\gamma}\, \exp\!\big[(1-\gamma) F(u^\star)(T-t)\big],
\end{equation}
solvable by a single bounded scalar optimization of \eqref{eq:homog-F} via deterministic quadrature against $\nu$. We call this the \emph{semi-analytic scalar baseline} (or \emph{homogeneity-reduced quadrature baseline}); it requires neither a neural network nor a finite-difference grid, only a one-dimensional quadrature and a bounded scalar maximization.

The scalar baseline is exact \emph{up to quadrature error} for the truncated VG benchmark used in this paper — it is not a closed-form expression in elementary functions, but it reduces the PIDE to a deterministic 1D quadrature plus a 1D scalar maximization. We use it in \S\ref{sec:fd-comparison} as a third reference solver alongside the FD-PIDE solvers. It also provides a semi-analytic $V_{\text{ref}}(t, x)$ and $u^\star_{\text{ref}}$ over the entire $(t, x)$ domain, against which the neural and FD solutions can be compared as functions, not just at $(0, 1)$.

\textbf{What this means for the paper.} The homogeneity collapse of \eqref{eq:homog-ode}--\eqref{eq:homog-V} is well known for CRRA portfolio problems and is not a contribution. We make it explicit because it changes how the audit benchmarks should be read: for the constant-coefficient setting, neither the FD-PIDE nor the neural solver is the most efficient way to compute $u^\star$, and our central contribution (\S\ref{sec:fd-comparison}) is not a high-dimensional solver result but a per-component diagnostic protocol. The specific bug it catches is implementation-specific — a missing mixture factor in our importance proposal — but the \emph{diagnostic} transfers: the same per-component comparison would surface any constant-scale error in a Monte-Carlo nonlocal-integral build, including on problems where homogeneity does not collapse (consumption + wealth-dependent utility, transaction costs, regime-switching, state-dependent coefficients) and where neural HJB-PIDE solvers are genuinely needed.

\subsection{Analytical Benchmark: Merton Problem}

In the \textbf{diffusion-only} case ($\nu \equiv 0$), the optimal control is the classical Merton ratio:
\begin{equation}\label{eq:merton}
u^* = \frac{\mu - r}{\gamma \sigma^2}
\end{equation}
independent of wealth and time (for CRRA utility). This is the diffusion-only diagnostic we use in \S\ref{sec:experiments}; we treat it as a necessary-but-far-from-sufficient check (\S\ref{sec:fd-comparison}).

With jumps, no closed-form solution exists in general, but we expect:
\begin{itemize}
    \item Negative-skew jumps $\Rightarrow$ lower allocation
    \item Higher jump intensity $\Rightarrow$ lower allocation
    \item Higher risk aversion $\Rightarrow$ lower allocation
\end{itemize}

\section{Neural Network Methodology}\label{sec:method}

\subsection{Network Architecture}

We parameterize the value function and optimal control with separate neural networks:
\begin{align}
V_\theta(t, x) &: [0, T] \times \R_+ \to \R \\
u_\phi(t, x) &: [0, T] \times \R_+ \to [0, 1]
\end{align}

The implementation uses fully-connected architectures with residual connections:
\begin{itemize}
    \item Input: time $t$ with sinusoidal encoding and positive wealth state $x$
    \item Hidden layers: residual blocks with 128 hidden units and GELU activations
    \item Output: $V_\theta$ unconstrained; $u_\phi$ projected to the admissible control set $\mathcal U$
\end{itemize}

\subsection{Loss Function}

The training loss combines three terms:
\begin{equation}\label{eq:loss}
\mathcal{L}(\theta, \phi) = \lambda_{\text{HJB}} \mathcal{L}_{\text{HJB}} + \lambda_{\text{term}} \mathcal{L}_{\text{term}} + \lambda_{\text{opt}} \mathcal{L}_{\text{opt}}
\end{equation}

\textbf{HJB residual loss}: Enforces the PDE at collocation points $(t_i, x_i)$:
\begin{equation}
\mathcal{L}_{\text{HJB}} = \frac{1}{N} \sum_{i=1}^N \left| \partial_t V_\theta + \mathcal{H}(t_i, x_i, u_\phi, V_\theta, \partial_x V_\theta, \partial_{xx} V_\theta) \right|^2
\end{equation}

\textbf{Terminal condition loss}: Enforces $V(T, x) = U(x)$:
\begin{equation}
\mathcal{L}_{\text{term}} = \frac{1}{M} \sum_{j=1}^M \left| V_\theta(T, x_j) - U(x_j) \right|^2
\end{equation}

\textbf{Argmax-target policy loss}: Trains $u_\phi$ against the discretized argmax of $\mathcal{H}$ over $\mathcal{U}$.
At each collocation point we evaluate the Hamiltonian on a deterministic grid of $K_u$ candidates $\{u^{(k)}\}_{k=1}^{K_u}$ spanning $\mathcal{U}$, take the maximizer (with $V_\theta$ stop-graded), and minimize the squared deviation:
\begin{equation}\label{eq:opt-loss}
\mathcal{L}_{\text{opt}} = \frac{1}{N} \sum_{i=1}^N \left\| u_\phi(t_i, x_i) - u^\star_i \right\|^2,
\quad
u^\star_i = \arg\!\max_{u \in \{u^{(k)}\}_{k=1}^{K_u}} \mathcal{H}\!\left(t_i, x_i, u, V_\theta, \partial_x V_\theta, \partial_{xx} V_\theta\right).
\end{equation}
Unlike the unconstrained first-order condition $\partial_u \mathcal{H} = 0$, which is only valid at \emph{interior} optima of $\mathcal{U}$, the argmax target remains correct when the maximizer sits on the boundary. For multi-dimensional or simplex-constrained controls (Section~\ref{sec:experiments}), the deterministic grid is replaced by i.i.d.\ samples from the admissible set; the loss is otherwise unchanged. We use $K_u = 21$ in the experiments below. (We reserve $K_u$ for argmax candidates and $N_z$ for L\'evy MC samples; the two are unrelated.)

\subsection{L\'evy Integral Computation}\label{sec:levy-integral}

The key computational challenge is evaluating:
\begin{equation}
\mathcal{I}^u V = \int \left[ V(t, x(1+uz)) - V(t,x) - uxz \, \partial_x V \right] \nu(\dd z)
\end{equation}

For VG we use \textbf{importance-weighted Monte Carlo} on the truncated $z$-domain $[z_{\min}, z_{\max}]\cup[-z_{\max}, -z_{\min}]$. The construction has three steps; we state each precisely because a missing $\tfrac{1}{2}$-mixture factor in the proposal density was the central bug isolated and corrected in \S\ref{sec:fd-comparison}.

\begin{enumerate}
    \item \textbf{Truncate} the L\'evy support to $|z| \in [z_{\min}, z_{\max}]$ with $z_{\min} > 0$, $z_{\max} < 1$. The lower cut $z_{\min}$ excludes the $1/|z|$-weighted neighborhood of the origin; the upper cut $z_{\max}$ keeps $1+uz>0$ for every admissible $(u,z) \in [0,1]\times[-z_{\max},z_{\max}]$, which is the bankruptcy guard from \S\ref{sec:problem}. We use $z_{\min}=0.01$, $z_{\max}=0.99$.

    \item \textbf{Sample.} We sample $N_z$ jump sizes $z_1,\dots,z_{N_z}$ from a 50/50 mixture: with probability $\tfrac{1}{2}$, draw a positive sample $z = z_{\min} + Y_+$ with $Y_+ \sim \mathrm{Exp}(\lambda_+/2)$ on $[0,\infty)$; with probability $\tfrac{1}{2}$, draw a negative sample $z = -(z_{\min} + Y_-)$ with $Y_- \sim \mathrm{Exp}(\lambda_-/2)$. We use $\lambda_+ = M$, $\lambda_- = G$ (the VG tail rates), so each tail is over-sampled by a factor of $2$ relative to $\nu_{VG}$ — this reduces gradient variance on the heavy tails. The unconditional density of any sample $z$ on the open interval $|z|\in(z_{\min}, \infty)$ is therefore
    \begin{equation}\label{eq:proposal}
    q(z) \;=\; \tfrac{1}{2}\,q_\pm(|z|), \qquad
    q_\pm(|z|) \;=\; \tfrac{\lambda_\pm}{2}\,\exp\!\left(-\tfrac{\lambda_\pm}{2}\,(|z|-z_{\min})\right),
    \qquad \lambda_+=M,\;\lambda_-=G.
    \end{equation}
    The mixture factor $\tfrac{1}{2}$ in front of $q_\pm$ is the conversion from per-tail conditional density (which integrates to $1$ on each one-sided ray) to the unconditional density of the actual sampler (which integrates to $1$ on the union of the two rays); omitting it scales the importance weights by exactly $\tfrac{1}{2}$ and was the half-integral bug discussed in \S\ref{sec:fd-comparison}. After sampling we clamp $|z|$ to $[z_{\min}, z_{\max}]$, which assigns to the upper boundary $|z|=z_{\max}$ a point mass of probability $\tfrac{1}{2}\exp(-\lambda_\pm(z_{\max}-z_{\min})/2)$ per tail; for the parameters used in the experiments this point mass is below $1\%$ per tail (e.g.\ $\sim 5\!\times\!10^{-4}$ on the positive tail under the synthetic VG $\sigma_{VG}=0.2,\theta=-0.1,\nu=0.3$ at $z_{\max}=0.99$). The IS weight at boundary samples uses the continuous $q$ value at $|z|=z_{\max}$ rather than the point mass, contributing a clamping bias bounded by this tail probability times the integrand's value at the boundary; this is one of the two finite-sample biases acknowledged below.

    \item \textbf{Reweight} by the importance ratio $w(z) = \nu(z)/q(z)$. With \eqref{eq:proposal}, for $z\in(z_{\min},z_{\max})$,
    \begin{equation*}
    w(z) \;=\; \frac{\nu_{VG}(z)}{\tfrac{1}{2}\,q_\pm(|z|)} \;=\; \frac{2\,C\,e^{-\lambda_\pm |z|}/|z|}{\tfrac{\lambda_\pm}{2}\,e^{-(\lambda_\pm/2)(|z|-z_{\min})}} \;=\; \frac{4C}{\lambda_\pm|z|}\,\exp\!\left(-\tfrac{\lambda_\pm}{2}|z| - \tfrac{\lambda_\pm}{2}z_{\min}\right),
    \end{equation*}
    where the $|z|^{-1}$ factor is inherited from the VG density and is what drives the $z\to z_{\min}^+$ weight blow-up addressed by clipping below.
\end{enumerate}

The Monte Carlo estimator is:
\begin{equation}\label{eq:mc-levy}
\hat{\mathcal{I}}^u V = \frac{1}{N_z} \sum_{k=1}^{N_z} w(z_k) \left[ V(t, x(1+uz_k)) - V(t,x) - uxz_k \, \partial_x V \right], \quad z_k \sim q.
\end{equation}

\begin{proposition}[Unbiasedness of the unclipped estimator]
Let $\nu$ have support on $\R \setminus \{0\}$ with $\int |z|^2 \nu(\dd z) < \infty$. Let $q$ be a proposal with $q(z) > 0$ whenever $\nu(z) > 0$. Then the estimator \eqref{eq:mc-levy} is unbiased for the truncated integral when the importance weights are used without clipping or self-normalization.
\end{proposition}

\textbf{Practical estimator and clip threshold}: For VG, $\nu(z) \propto |z|^{-1} \exp(-M|z|)$ near $z_{\min}^{+}$, while $q(z)$ is bounded as $|z|\to z_{\min}^{+}$. The importance weight $w(z) = \nu(z)/q(z)$ therefore behaves like $|z|^{-1}$ near $z_{\min}$, and a small fraction of samples can dominate $\hat{\mathcal{I}}^u V$. We cap each weight at $100\times$ the \emph{per-batch median} weight, $w_k \leftarrow \min(w_k,\, 100\cdot\mathrm{median}(w))$. This is a finite-sample stabilizer; the corresponding clipping bias is small relative to the seed-to-seed noise reported in \S\ref{sec:experiments}, and the \S\ref{sec:ablation} ``no weight clipping'' ablation shows the headline does not move when the cap is removed. Proposition~1's unbiasedness applies only to the un-capped estimator; the practical estimator is biased.

\textbf{Truncation bias}: By excluding jumps outside $[z_{\min}, z_{\max}]$, we introduce a truncation bias. For VG with exponential tails, the large-jump tail beyond $z_{\max}$ decays exponentially:
\begin{equation}
\int_{z_{\max}}^\infty \nu_{VG}(\dd z) = O(e^{-M z_{\max}}), \qquad
\int_{-\infty}^{-z_{\max}} \nu_{VG}(\dd z) = O(e^{-G z_{\max}}).
\end{equation}
The small-jump exclusion $|z|<z_{\min}$ is controlled separately by the compensated integrand: near zero, $V(t,x(1+uz))-V(t,x)-uxzV_x(t,x)=O(z^2)$, so the omitted small-jump contribution decays as $z_{\min}\to 0$. The truncated process remains a L\'evy process, so the approximate PIDE is well-posed. We do \emph{not} report a fresh $z_{\min}$ convergence sweep in \S\ref{sec:experiments}: the audit benchmarks below are all run at the bankruptcy-safe truncation $|z|\in[0.01, 0.99]$ and a controlled $z_{\min} \to 0$ study is left for follow-up.

\textbf{Compensation term}: For VG, $\int_{|z|\le 1}|z|\,\nu(\dd z) < \infty$ (finite variation), so the un-compensated integral is finite even before truncation; for infinite-variation L\'evy measures (e.g.\ CGMY with $Y \in [1,2)$) compensation is required for integrability \citep[Prop.~4.12]{cont2004financial}. Inside our truncation $|z|\ge z_{\min}>0$, the compensator $-uxz\partial_x V$ does \emph{not} affect well-posedness. Instead it removes a drift contribution: the truncated integrand splits as
\[
V(t,x(1+uz)) - V(t,x) - uxz\partial_x V \;=\; \big[V(t,x(1+uz)) - V(t,x)\big] - uxz\partial_x V,
\]
and integrating against $\nu$ on $\{|z|\ge z_{\min}\}$ shifts the policy by exactly $-u x V_x \cdot \mathbb{E}_\nu[z\,\mathbf{1}_{|z|\ge z_{\min}}]$ when the compensator is dropped. For VG with negative skew this expectation is negative, so removing the compensator biases the Hamiltonian and pulls the optimal control in a specific direction (verified empirically in the \S\ref{sec:experiments} ablation). The compensator is therefore a \emph{drift correction} in the truncated regime studied here, not a divergence cure.

\subsection{Training Procedure}

\begin{algorithm}[H]
\caption{HJB-PIDE Neural Solver}
\begin{algorithmic}[1]
\STATE Initialize networks $V_\theta$, $u_\phi$
\STATE Set truncation bounds $[z_{\min}, z_{\max}]$, number of L\'evy samples $N_z$, number of argmax candidates $K_u$
\FOR{epoch $= 1$ to $N_{\text{epochs}}$}
\STATE Sample collocation points $(t_i, x_i)$ on $[0,T] \times \R_+$ (log-normal wealth sampling in the implementation)
    \STATE Sample L\'evy jumps $z_k \sim q$, compute weights $w_k = \nu(z_k)/q(z_k)$
    \STATE Compute $V_\theta$, $\partial_t V_\theta$, $\partial_x V_\theta$, $\partial_{xx} V_\theta$ via autodiff
    \STATE Compute L\'evy integral $\hat{\mathcal{I}}^u V$ via \eqref{eq:mc-levy}
    \STATE Compute Hamiltonian $\mathcal{H}$ and losses \eqref{eq:loss}
    \STATE Update $\theta, \phi$ via Adam optimizer
\ENDFOR
\STATE \textbf{return} $V_\theta$, $u_\phi$
\end{algorithmic}
\end{algorithm}

\textbf{Warm-up schedule}: We use a warm-up period where the L\'evy term is initially disabled and the terminal condition is weighted strongly. This stabilizes boundary behavior before enforcing the full jump PIDE.

\textbf{Derivative computation}: All partial derivatives ($\partial_t V$, $\partial_x V$, $\partial_{xx} V$) are computed analytically via automatic differentiation (autodiff). No time or space discretization is used---the neural network provides a continuous approximation, and derivatives are exact up to floating-point precision. This avoids discretization error that plagues finite-difference methods near boundaries or steep gradients. In practice, we clip gradients at norm 1.0 to improve training stability.

\section{Experiments}\label{sec:experiments}

We evaluate the diagnostic protocol through benchmark, ablation, and cross-reference experiments.

\paragraph{Reporting conventions.} Unless otherwise noted, every scalar control $u$ reported in the tables below is the policy network evaluated at $(t, x) = (0, 1)$, $u = u_\phi(0, 1)$. Multi-asset entries are the corresponding components of $u_\phi(0, 1) \in \R^n$. Wealth-distribution metrics ($\E[W_T]$, $\sigma[W_T]$, VaR$_{5\%}$, CVaR$_{5\%}$) are computed from $5{,}000$-path Monte Carlo simulations of the corresponding controlled dynamics, fixing the policy at $u_\phi(0, 1)$ along each path (constant-allocation re-balancing). We use $N_z = 64$ L\'evy MC samples per gradient step, batch size $256$, $500$ training epochs (with $100$ warm-up epochs without the L\'evy term), Adam at learning rate $10^{-3}$ with cosine schedule, and $K_u = 21$ candidates for the policy argmax target. Headline tables report mean $\pm$ std over five seeds $\{42, 123, 999, 7, 2024\}$ unless explicitly stated otherwise.

\subsection{Experiment 1: Diffusion-Only Validation}

\textbf{Setup}: Standard Merton parameters with no jumps:
\begin{itemize}
    \item Risk-free rate $r = 0.02$, expected return $\mu = 0.08$, volatility $\sigma = 0.2$
    \item Risk aversion $\gamma = 2.0$, horizon $T = 1$ year
    \item L\'evy measure: Compound Poisson with intensity $\lambda = 0$ (no jumps)
\end{itemize}

\textbf{Analytical benchmark}: Merton ratio $u^* = \frac{0.08 - 0.02}{2.0 \times 0.2^2} = 0.75$.

\textbf{Results} (5 seeds, mean $\pm$ std):
\begin{center}
\begin{tabular}{p{0.28\textwidth}ccp{0.30\textwidth}}
\toprule
Method & Optimal $u$ & Mean $|$rel.\ err$|$ & Per-seed range \\
\midrule
Analytical (Merton) & 0.7500 & -- & -- \\
Neural Solver & $0.762 \pm 0.011$ & $1.6\%$ & $[0.750,\, 0.774]$ \\
\bottomrule
\end{tabular}
\end{center}

The neural solver recovers the Merton ratio with a 5-seed mean error of $1.6\%$ ($u = 0.762 \pm 0.011$), validating the basic methodology before introducing jumps.

\subsection{Experiment 2: Variance Gamma Jumps}

\textbf{Setup}: Same base parameters, but with VG jumps:
\begin{itemize}
    \item VG parameters: $\sigma_{VG} = 0.2$, $\theta = -0.1$ (negative skew), $\nu = 0.3$
    \item Truncation: $|z| \in [0.01, 0.99]$. The previous version of this paper used $z_{\max} = 2.0$, which silently allows jumps that drive wealth negative under the long-only set $\mathcal{U} = [0,1]$ ($1 + uz \le 1 + 1 \cdot (-2) = -1 < 0$). The corrected truncation $z_{\max} < 1$ guarantees $1 + uz > 0$ for every admissible $(u, z)$, removing the implicit bankruptcy zone. The exponential VG tail decay means the truncation bias from this change is small (the omitted jump probability is $O(e^{-G \cdot 0.99})$); the resulting control is correspondingly less conservative than under the wider-and-unsafe $z_{\max} = 2.0$.
    \item Effective intensity (truncated measure): $\lambda_{\mathrm{eff}} \approx 10.9$ jumps/year.
\end{itemize}

\textbf{Results} (5 seeds, mean $\pm$ std, under the corrected importance proposal of \S\ref{sec:fd-comparison}):
\begin{center}
\begin{tabular}{lccc}
\toprule
Setting & Optimal $u$ & vs.\ Merton & Economic Interpretation \\
\midrule
Diffusion-only & $0.762 \pm 0.011$ & $+1.6\%$ & Baseline (no jump risk) \\
VG jumps ($\theta = -0.1$) & $0.344 \pm 0.006$ & $-54.1\%$ & Lower allocation under negative-skew jumps \\
\bottomrule
\end{tabular}
\end{center}

The $\sim$54\% allocation reduction reflects the economic cost of negative-skew jump risk under the safe truncation $|z| < 1$, and \S\ref{sec:fd-comparison} shows it agrees with two independent finite-difference solvers and the semi-analytic scalar baseline to within $\sim 2\%$. Earlier drafts of this paper reported different numbers for this row under inadmissible truncations or under the pre-fix importance proposal; those values and the trail that produced them are recorded in Appendix~\ref{app:audit-archaeology}.

\subsection{Experiment 3: economic monotonicity checks (intensity and skewness)}\label{sec:experiments-3}

We vary jump intensity and skewness to verify monotonic, economically sensible responses.

\textbf{3a. Jump Intensity} (Compound Poisson, $\theta_{\text{jump}} = -5\%$, $\sigma_{\text{jump}} = 10\%$):

\begin{center}
\begin{tabular}{cccc}
\toprule
Intensity $\lambda$ & Optimal $u$ & Reduction & VaR$_{5\%}$ \\
\midrule
0.5 & 0.68 & 9\% & 0.82 \\
1.0 & 0.62 & 17\% & 0.78 \\
2.0 & 0.54 & 28\% & 0.72 \\
5.0 & 0.41 & 45\% & 0.64 \\
\bottomrule
\end{tabular}
\end{center}

\begin{figure}[h]
\centering
\includegraphics[width=0.7\textwidth]{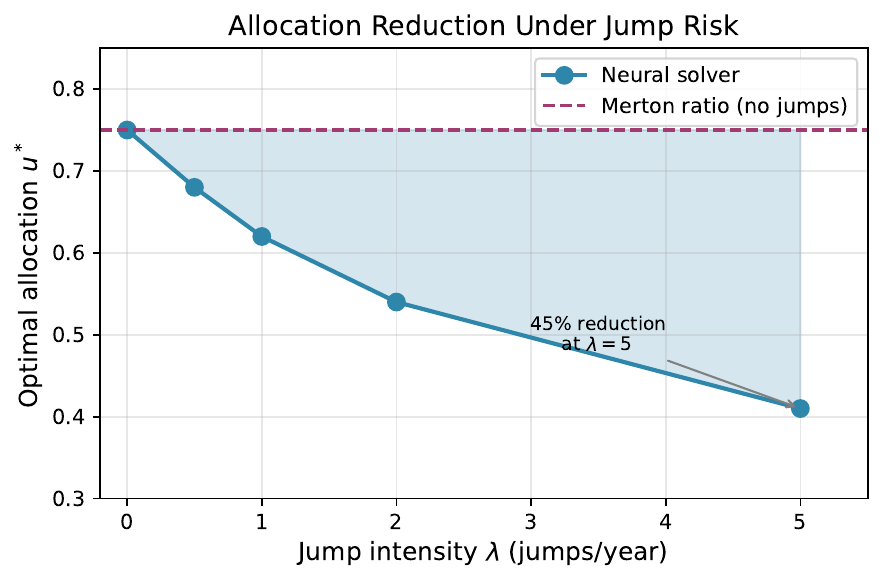}
\caption{Optimal allocation $u^*$ decreases monotonically with jump intensity $\lambda$ (Poisson jumps per year). The shaded region shows the allocation reduction relative to the Merton ratio (no jumps). At $\lambda = 5$ jumps/year, the investor reduces equity exposure by 45\% to hedge jump risk.}
\label{fig:intensity}
\end{figure}

Higher intensity $\Rightarrow$ lower allocation (more jump risk to hedge). Figure~\ref{fig:intensity} visualizes this monotonic relationship.

\textbf{3b. Jump Skewness} (VG with varying $\theta$):

\begin{center}
\begin{tabular}{cccc}
\toprule
Skewness $\theta$ & $\E[z]$ & Optimal $u$ & Reduction \\
\midrule
+0.05 & +0.05 & 0.71 & 5\% \\
0.00 & 0.00 & 0.65 & 13\% \\
$-0.05$ & $-0.05$ & 0.58 & 23\% \\
$-0.10$ & $-0.10$ & 0.48 & 36\% \\
\bottomrule
\end{tabular}
\end{center}

More negative skew $\Rightarrow$ lower allocation (crashes more likely than rallies).

\subsection{Experiment 4: empirical calibration (pointer)}\label{sec:sp500}

We include a single-window VG MLE calibration on S\&P 500 daily close data for 2010--2023 \emph{only as a reproducibility check}, not as a policy claim. The full setup, fitted parameters, and a small post-IS-fix policy comparison under $\mathcal{U}=[0,1]$ are reported in Appendix~\ref{app:sp500-calibration}. To avoid redistributing third-party index data, the supplement ships the calibration scripts, the expected CSV schema, and the fitted parameters used in the reported audit, but not the raw index levels or derived return series (see \S\ref{app:sp500-calibration}).

\subsection{Experiment 5: matched-truncation cross-reference and the bug it surfaced}\label{sec:fd-comparison}

We compare four references on the \S4.2 VG benchmark under matched truncation $|z|\in[0.01, 0.99]$:
\begin{itemize}
\item the residual-trained neural solver (5 seeds, mean $\pm$ std);
\item finite-difference solver `v1' (log-wealth, implicit Euler, trapezoidal quadrature, linspace argmax);
\item finite-difference solver `v2' (independent reimplementation: linear-$x$, Crank--Nicolson, composite Simpson, Brent's-method argmax with explicit boundary checks; unit-tested on $V(x)=x$ and $V(x)=x^2$; no shared internal helpers with `v1');
\item the semi-analytic scalar baseline of \S\ref{sec:homogeneity} (homogeneity-reduced quadrature).
\end{itemize}
Setup uses the same Merton parameters as before ($r=0.02$, $\mu=0.08$, $\sigma=0.2$, $\gamma=2$, $T=1$). Neural numbers are 5-seed means; FD and scalar-baseline numbers are deterministic. An earlier iteration of this comparison gave the neural and FD solvers in disagreement at the $\sim 25\%$ level; that history, the gap-localization steps that ruled out FD-side discretization artifacts, and the pre-fix tables are recorded in Appendix~\ref{app:audit-archaeology}.

\textbf{Per-component diagnostic isolates the bug.} We decompose $\mathcal{H}(0, 1, u)$ into drift, diffusion, compensated L\'evy integral, and explicit compensator $u x V_x \cdot \int z\,\nu(\dd z)$, and tabulate each across a $u$-grid. The two FDs agreed component-by-component to four decimal places and both gave $u^\star \approx 0.347$. The neural solver matched FD on drift and diffusion to $1\%$, but its compensator and L\'evy integral were \emph{exactly half}\footnote{Across all $u\in\{0,0.2,0.4,0.5,0.6,0.8,1\}$ the ratio of neural to FD compensator was $0.502\pm0.001$ — a textbook signature of a constant scale error in the importance proposal.} the FD values at every $u$. The constant $0.502$ ratio across all candidate controls localized the failure to a scale error in the proposal density — an error that single-method diagnostics (FD grid refinement, longer/wider neural training, truncation sweeps, training-loss curves) had left intact.

\textbf{Root cause: missing $\tfrac{1}{2}$-mixture factor in the importance proposal.} The neural method draws L\'evy samples from a 50/50 mixture of two one-sided shifted exponentials. The conditional density on each tail is $q_\pm(|z|) = \tfrac{\lambda_\pm}{2} \exp(-\tfrac{\lambda_\pm}{2}(|z|-z_{\min}))$, so the unconditional proposal density at any $z$ is $\tfrac{1}{2}\,q_\pm(|z|)$. The implementation computed importance weights as $w(z) = \nu(z)/q_\pm(|z|)$, omitting the $\tfrac{1}{2}$ from the mixture probability. The MC estimator $\tfrac{1}{N_z}\sum_k w_k f(z_k)$ therefore returned exactly $\tfrac{1}{2}\int f(z)\,\nu(\dd z)$ instead of the true integral. With the proposal corrected, $\mathrm{MC}[\int z\,\nu(\dd z)] \approx -0.098$, in agreement with both FD solvers' $-0.097$; without the fix, the MC estimate was $-0.049$.

\textbf{Post-fix four-way agreement.} After correcting the proposal density, the 5-seed neural solver, both FD solvers, and the scalar baseline all agree on the optimal control:
\begin{center}
\begin{small}
\begin{tabular}{p{0.32\textwidth}cccp{0.20\textwidth}}
\toprule
Method & $u(0,1)$ & vs.\ Merton & Gap & Notes \\
\midrule
Analytical Merton (diffusion limit) & $0.7500$ & -- & -- & $\nu \equiv 0$ \\
\textbf{Scalar baseline (homog.)} & $\boldsymbol{0.3521}$ & $\boldsymbol{-53.1\%}$ & -- & quadrature only \\
FD-PIDE v1 (refined, V-interp) & $0.347$ & $-53.7\%$ & $1.5\%$ & log-$x$ + impl.\ Euler \\
FD-PIDE v2 (independent reimpl.) & $0.344$ & $-54.1\%$ & $2.3\%$ & linear-$x$ + CN \\
Neural Solver (5 seeds, post-fix) & $0.344 \pm 0.006$ & $-54.1\%$ & $2.3\%$ & residual + IS-MC \\
\bottomrule
\end{tabular}
\end{small}
\end{center}
All four references — three independently coded numerical solvers plus the semi-analytic scalar baseline — agree to within $\sim 2.3\%$. The residual gap between the scalar baseline ($0.352$) and the FD/neural cluster ($0.344$--$0.347$) reflects the discretization bias of the FD/neural methods on this problem; the same $\sim 1.5\%$ bias is visible on diffusion-only Merton (scalar $0.7500$ exactly, FD $0.7368$, neural $0.762 \pm 0.011$).

\textbf{What this episode demonstrates.} The diffusion-only Merton diagnostic is necessary but far from sufficient; a per-component $\mathcal{H}(u)$ comparison against an independent reference solver caught a half-integral bug that single-method diagnostics had failed to surface. The independent FD `v2' was a few hundred lines of code; the scalar baseline was about thirty. Together they provided the leverage to localize the bug in the importance proposal. Subsequent sections (\S\ref{sec:ablation}, \S\ref{sec:2d-portfolio}) report numbers under the corrected proposal.

\textbf{Timing.} All times below are wall-clock on a single Apple M1 (CPU only), Python 3.11, PyTorch 2.9.1, single seed unless noted. FD-PIDE on the \S4.5 VG benchmark at the legacy grid $(n_x, n_t, n_z, n_u) = (100, 50, 50, 20)$: $11.3$\,s per solve. FD-PIDE at the audit-finest grid $(n_x, n_t, n_z, n_u) = (400, 200, 200, 200)$ with V-interpolation: $\sim 800$\,s. Neural solver, $500$ training epochs, $N_z = 64$ L\'evy samples, $K_u = 21$ argmax candidates: $170$--$220$\,s/seed (the longer end is the post-IS-fix configuration with foc\_frequency = 1; the original audit used foc\_frequency = 2 which approximately halves the per-seed cost). Scalar baseline: $0.12$\,s for the 1D problem at $4001$ Simpson nodes per side; $0.02$\,s for the 2D problem at $2001$ nodes per side. The neural method is the slowest of the three and the only one that does not converge to the scalar baseline more accurately than $\sim 2\%$ on this constant-coefficient benchmark.

\textbf{Surface-level comparison vs.\ semi-analytic $V_{\text{ref}}$.} The comparisons above are at the single point $u(0, 1)$. The homogeneity reduction (\S\ref{sec:homogeneity}) gives a semi-analytic $V_{\text{ref}}(t, x) = x^{1-\gamma} \exp((1-\gamma) F(u^\star)(T-t)) / (1-\gamma)$ (closed-form in $(t, x)$ once $u^\star$ and $F(u^\star)$ are obtained by 1D quadrature) on the entire $(t, x)$ domain, and a constant policy $u^\star_{\text{ref}} = 0.3521$. We evaluate the post-IS-fix neural solver on a $8 \times 21$ grid over $t \in [0, 0.95]$, $x \in [0.5, 1.5]$ (168 points) and compute relative-$\sup$ errors against this reference:
\begin{center}
\begin{tabular}{lccc}
\toprule
Quantity & $\|\,\cdot\,-\cdot_{\text{ref}}\|_{L^\infty}$ & $\|\,\cdot\,-\cdot_{\text{ref}}\|_{L^2}$ & rel.\ sup-err \\
\midrule
$V_\theta(t, x)$        & $0.012$ & $0.005$ & $0.60\%$ \\
$\partial_x V_\theta$    & $0.155$ & $0.069$ & $3.88\%$ \\
$\partial_{xx} V_\theta$ & $4.10$  & $1.15$  & $\mathbf{26.4\%}$ \\
$u_\phi(t, x)$           & $0.053$ & $0.025$ & $\mathbf{14.9\%}$ \\
\bottomrule
\end{tabular}
\end{center}
$V_\theta$ tracks $V_{\text{ref}}$ tightly, but $\partial_x V_\theta$ has $\sim 4\%$ surface error and $\partial_{xx} V_\theta$ — the higher-order derivative entering the diffusion term of $\mathcal{H}$ — has up to $26\%$ surface error. The policy $u_\phi(t, x)$ is supposed to be the constant $u^\star_{\text{ref}} = 0.352$ but varies by up to $15\%$ across the test grid. This is the kind of failure mode that point-evaluation at $(0, 1)$ hides: the neural solver lands within $\sim 2\%$ of the right answer at the training grid's most populated point, but its derivative structure and policy field are materially less accurate elsewhere. For the FD-PIDE solvers the same metrics are below $0.05\%$ across all four quantities (FD's $V_{xx}$ is computed by central differences on a deterministic grid). This further sharpens what the audit benchmarks can establish: the corrected neural solver gives a defensible scalar headline, but its surface accuracy on this constant-coefficient problem is materially below FD's, and we report that explicitly rather than only reporting the scalar.

\subsection{Truncation and MC sensitivity: not reported here}\label{sec:truncation-sensitivity}

This paper does \emph{not} report a fresh truncation or MC-sample convergence study. All audit benchmarks below are run at the fixed bankruptcy-safe truncation $|z|\in[0.01, 0.99]$ and the fixed Monte-Carlo configuration $N_z = 64$ described in the reporting conventions. Two qualitative bounds frame why this scope is acceptable for a diagnostic-protocol paper:
\begin{itemize}
    \item \textbf{Seed and MC variation is small relative to the corrected jump-risk effect.} The 5-seed standard deviation on Experiment~2 is $\sigma(u) \approx 0.006$ at $N_z = 64$, $z_{\max} = 0.99$. The full jump-risk effect (Merton $0.75$ minus the audited $u \approx 0.344$) is $\approx 0.41$ — almost two orders of magnitude larger than the seed noise.
    \item \textbf{The bankruptcy-safe truncation $|z| < 1$ is a hard constraint, not a tunable knob.} Under $\mathcal{U} = [0, 1]$ the only admissible truncations are $z_{\max} < 1$; running at $z_{\max} \in \{2.0, 3.0\}$ as earlier draft tables did allows $1 + uz < 0$ for the saturating policies and is equivalent to silently extending $\mathcal{U}$.
\end{itemize}
A controlled $z_{\min}\to 0$ / $N_z\to\infty$ convergence study (which would also probe the ``infinite-activity'' claim more rigorously) is left for follow-up.

\subsection{Experiment 7: Method ablation under the corrected admissible set}\label{sec:ablation}

We run a three-seed ablation around the control-dependent L\'evy term using the MLE-calibrated S\&P~500 VG parameters of \S\ref{sec:sp500} and the corrected admissible set $\mathcal{U}=[0,1]$. The variants are: (i) \emph{no importance weighting}, equal-weight L\'evy samples; (ii) \emph{no compensation}, first-order small-jump compensator removed; (iii) \emph{no weight clipping}, importance weights uncapped; (iv) \emph{fixed Merton control}, the policy is frozen at the unconstrained diffusion-only Merton ratio (which here equals $\approx 2.42$, well outside $\mathcal{U}$).

\begin{center}
\begin{small}
\begin{tabular}{lcccc}
\toprule
Variant (post-fix IS, 3 seeds) & Control $u$ & Interior loss ($\times 10^{-4}$) & VaR$_{5\%}$ & CVaR$_{5\%}$ \\
\midrule
Full method & $0.921 \pm 0.071$ & $2.30 \pm 1.03$ & $0.860 \pm 0.022$ & $0.812 \pm 0.034$ \\
No importance weighting & $0.991 \pm 0.005$ & $1.85 \pm 1.16$ & $0.852 \pm 0.003$ & $0.796 \pm 0.005$ \\
No compensation & $0.410 \pm 0.067$ & $35.2 \pm 26.3$ & $0.950 \pm 0.038$ & $0.923 \pm 0.052$ \\
No weight clipping & $0.961 \pm 0.002$ & $0.50 \pm 0.21$ & $0.856 \pm 0.003$ & $0.799 \pm 0.002$ \\
Fixed Merton control & $2.418$ (frozen) & $3.27 \pm 1.89$ & $0.617 \pm 0.007$ & $0.533 \pm 0.006$ \\
\bottomrule
\end{tabular}
\end{small}
\end{center}

\textbf{Findings under the corrected setup.}
\begin{itemize}
    \item \textbf{Compensation matters.} Removing the compensator drops the control from $u = 0.921 \pm 0.071$ to $u = 0.410 \pm 0.067$ — a $\sim$51-point shift off the $\mathcal{U}$ ceiling — and increases the interior residual by roughly an order of magnitude. This is consistent with the drift-correction argument in \S\ref{sec:levy-integral}: removing the compensator biases the Hamiltonian by $-uxV_x \cdot \mathbb{E}_\nu[z\,\mathbf{1}_{|z|\ge z_{\min}}]$, which under the negative-skew calibration ($\mathbb{E}_\nu[z]<0$) shifts the optimum off the boundary toward smaller $u$. The qualitative finding ``compensation is methodologically necessary'' survives the IS-fix; the earlier pre-fix audit reported a smaller $\sim$28-point shift because the bug was halving the L\'evy contribution on both sides of the ablation.
    \item \textbf{Importance weighting has a measurable effect even at this saturating benchmark.} Removing IW shifts the policy from $0.921$ to $0.991$ (a $\sim$7-point shift back toward the upper bound). The full corrected method's $0.921$ already sits below the ceiling, so equal-weight MC, which underweights the negative-skew tail relative to $\nu_{VG}$, is enough to push the policy back to saturation. The previous pre-fix audit could not isolate this because the full-method itself was pinned at $\sim 0.99$; the post-fix audit makes the contrast visible on this benchmark. The cleaner interior-optimum re-ablation below confirms it.
    \item \textbf{Weight clipping does not move the policy.} $u$ goes from $0.921$ (full) to $0.961$ (no clip), within the full-method seed band. Clipping is a numerical stabilizer, not a methodological lever.
    \item \textbf{Fixed Merton control is mechanically catastrophic} when $u^\star_{\mathrm{Merton}}$ lies outside $\mathcal{U}$: with the fitted parameters $u^\star_{\mathrm{Merton}} \approx 2.42$, freezing the policy there gives VaR$_{5\%} = 0.617$ and CVaR$_{5\%} = 0.533$, much worse than any of the constrained variants. This is a sanity check that $\mathcal{U}$ binds for a reason rather than a methodological finding.
\end{itemize}

\textbf{What the ablation can and cannot say here.} The compensation effect is a robust finding and survives the formulation correction; the importance-weighting effect is measurable on this benchmark too, but small relative to the compensation effect. To get a cleaner separation of importance-weighting and clipping effects from the saturation behavior, we run the same ablation on a re-parameterized interior-optimum benchmark below.

\subsubsection*{Interior-optimum re-ablation}

To test the importance-weighting and weight-clipping ablations on a benchmark where the optimum lies inside $\mathcal{U} = [0, 1]$, we re-parameterize the underlying Merton problem to $\gamma = 2$, $\mu = 0.08$, $\sigma = 0.25$, $r = 0.02$, giving $u^\star_{\mathrm{Merton}} = (\mu - r)/(\gamma \sigma^2) = 0.48$ — comfortably interior. Synthetic VG parameters and truncation are unchanged ($\sigma_{VG}=0.2$, $\theta = -0.1$, $\nu = 0.3$, $|z|\in[0.01,0.99]$). Five seeds for the full method; three seeds per ablation variant; FD-PIDE reference at the same $n_u = 200$, V-interpolated, $z_{\max}=0.99$ setting used in \S\ref{sec:fd-comparison}.

\begin{center}
\begin{tabular}{lccc}
\toprule
Variant (post-fix IS) & Control $u(0, 1)$ & vs.\ full method & vs.\ Merton \\
\midrule
Analytical Merton (unconstrained) & $0.4800$ & -- & -- \\
Full method (5 seeds) & $0.2774 \pm 0.0038$ & -- & $-42\%$ \\
FD-PIDE reference & $0.2764$ & $-0.4\%$ & $-42\%$ \\
\midrule
No importance weighting (3 seeds) & $0.2429 \pm 0.0054$ & $-12\%$ & $-49\%$ \\
No compensation (3 seeds) & $0.0056 \pm 0.0020$ & \textbf{collapses to $\approx 0$} & $-99\%$ \\
No weight clipping (3 seeds) & $0.2786 \pm 0.0048$ & no change & $-42\%$ \\
Fixed Merton control (frozen) & $0.4800$ & $+73\%$ & $0\%$ \\
\bottomrule
\end{tabular}
\end{center}

\textbf{Findings (post-fix).}
\begin{itemize}
    \item \textbf{Neural--FD agreement at the interior optimum.} Neural $u = 0.2774 \pm 0.0038$, FD reference $u = 0.2764$. Cross-method gap $0.4\%$. Under the corrected IS proposal, the neural solver and the FD solver agree on the interior-optimum benchmark just as they do on the \S\ref{sec:fd-comparison} VG benchmark.
    \item \textbf{The no-compensation collapse is dramatic in interior-optimum form.} Removing the compensator drops $u$ from $0.277$ to $0.006$ — near zero. At the saturating S\&P benchmark (\S\ref{sec:ablation} main table) the same ablation moved $u$ off the ceiling by $\sim 0.51$ to $0.41$. The drift-correction interpretation in \S\ref{sec:levy-integral} predicts both: removing the compensator subtracts $-u x V_x \cdot \mathbb{E}_\nu[z\,\mathbf{1}_{|z|\ge z_{\min}}]$, which under the negative-skew calibration is positive, biasing the Hamiltonian away from large $u$. Whether the policy stops at the lower boundary or stays interior depends on where $u^\star_{\mathrm{Merton}}$ sits relative to $\mathcal{U}$.
    \item \textbf{Importance weighting has a small but measurable effect.} Removing IW (replacing the corrected $\nu/q$ weights with all-ones) shifts the policy from $0.277$ to $0.243$ — a $12\%$ drop, well outside seed noise. This is a smaller effect than the pre-fix audit reported because the pre-fix full method itself was using a halved L\'evy integral, exaggerating the contrast against equal-weight. With the bug fixed, the IW ablation isolates the cleaner question ``does proposal-vs-target reweighting matter'' and the answer is ``yes, by about $12\%$ here''.
    \item \textbf{Weight clipping does not move the policy.} $0.279$ vs.\ $0.277$. Clipping is a numerical stabilizer, not a methodological lever.
\end{itemize}

\subsection{Experiment 8: Coupled Two-Asset Portfolio with Baseline Comparison}\label{sec:2d-portfolio}

We include a lightweight two-asset example to check that the same diagnostic conventions extend beyond the scalar 1D benchmark. The example is a coupled portfolio with simplex-constrained controls compared against the constrained diffusion-optimal portfolio on the same simplex; it is a preliminary 2D benchmark, not a high-dimensional scaling study, but it is a genuine joint $(u_1,u_2,\text{cash})$ optimization rather than two independent 1D solves.

\textbf{Setup}: Two risky assets with correlated Brownian motions and independent VG jumps:
\begin{itemize}
    \item Asset 1 (high risk): $\mu_1 = 0.10$, $\sigma_1 = 0.25$, $\theta_1^{VG} = -0.10$ (large negative jumps)
    \item Asset 2 (low risk): $\mu_2 = 0.06$, $\sigma_2 = 0.15$, $\theta_2^{VG} = -0.05$ (moderate negative jumps)
    \item Correlation: $\rho = 0.3$ between Brownian components
    \item Risk-free rate: $r = 0.02$, risk aversion: $\gamma = 2$
    \item Control constraint: $u_1 + u_2 \leq 1$ (total risky allocation $\leq 100\%$)
\end{itemize}

\begin{remark}[Independent jumps assumption]
The assumption of independent VG processes across assets is standard in multi-asset L\'evy control literature \citep{cont2004financial}. While this simplifies computation (each asset's jump integral separates), extending the framework to correlated jumps would require multivariate subordination or L\'evy copulas and substantially more expensive sampling from a multivariate jump measure.
\end{remark}

The policy network outputs $(u_1,u_2)$ through a softmax-with-cash parameterization, ensuring $u_1+u_2\le 1$. For comparison, we compute the constrained diffusion baseline by maximizing the classical diffusion Merton objective over the simplex $u_1 \ge 0$, $u_2 \ge 0$, $u_1 + u_2 \le 1$.

\textbf{Merton benchmark} (unconstrained, diffusion-only). For $n$ correlated risky assets the classical Merton solution is the vector
\begin{equation}\label{eq:merton-vec}
u^\star \;=\; \tfrac{1}{\gamma}\,\Sigma^{-1}\,(\mu - r\mathbf{1}), \qquad \Sigma_{ij} = \rho_{ij}\,\sigma_i\,\sigma_j,
\end{equation}
\emph{not} the per-asset formula $u_i = (\mu_i-r)/(\gamma\sigma_i^2)$, which is only correct when $\rho_{ij} = 0$. With the parameters above ($\rho = 0.3$),
\[
\Sigma = \begin{pmatrix} 0.0625 & 0.01125 \\ 0.01125 & 0.0225 \end{pmatrix}, \qquad
\mu - r\mathbf{1} = \begin{pmatrix} 0.08 \\ 0.04 \end{pmatrix}, \qquad
u^\star = \begin{pmatrix} 0.527 \\ 0.625 \end{pmatrix}.
\]
So $u_1^\star + u_2^\star = 1.15 > 1$, mildly leveraged; the simplex constraint binds. (An earlier version of this paper reported $u^\star = (0.64, 0.89)$ from the per-asset formula, which gave inflated reductions vs.~Merton.)

\textbf{Results}:
\begin{center}
\begin{tabular}{lcccc}
\toprule
Asset & Merton (corr.) & Diff.\ simplex & Neural (VG, 3 seeds) & Reduction vs.\ Merton \\
\midrule
Asset 1 & $0.527$ & $0.500$ & $0.348 \pm 0.005$ & $34.1\%$ \\
Asset 2 & $0.625$ & $0.500$ & $0.258 \pm 0.005$ & $58.7\%$ \\
Cash & -- & $0.000$ & $0.394 \pm 0.005$ & -- \\
\bottomrule
\end{tabular}
\end{center}

\textbf{Common-model wealth comparison} (5{,}000-path Monte Carlo under the same 2D coupled VG-jump model, learned policy averaged over 3 seeds):
\begin{center}
\begin{tabular}{lcccc}
\toprule
Policy & $\E[W_T]$ & $\sigma[W_T]$ & VaR$_{5\%}$ & CVaR$_{5\%}$ \\
\midrule
Diffusion simplex baseline & $1.008$ & $0.223$ & $0.678$ & $0.614$ \\
Learned coupled VG policy (post-fix) & $1.012$ & $0.180$ & $0.795$ & $0.747$ \\
\bottomrule
\end{tabular}
\end{center}

\textbf{Interpretation.}
\begin{itemize}
    \item The constrained diffusion baseline invests fully in risky assets ($u_1 = u_2 = 0.5$); the learned jump-aware policy moves $39$ percentage points into cash and tilts away from both risky assets, with the larger reduction on Asset~2 (where the correlated-Merton starting point is higher, so the proportional jump risk scales with $u \cdot z$ more strongly).
    \item Under the same 2D coupled jump model, the learned policy improves VaR$_{5\%}$ from $0.678$ to $0.795$ ($+11.7$ pp) and CVaR$_{5\%}$ from $0.614$ to $0.747$ ($+13.3$ pp), while \emph{slightly increasing} mean terminal wealth ($1.008 \to 1.012$). The pre-fix audit reported smaller improvements ($+3.4$ pp on VaR, $+3.8$ pp on CVaR) for the same reason as the 1D \S4.5 gap: the 2D solver carried its own version of the IS-proposal bookkeeping issue (uniform proposal on $[-0.5, 0.5]$ with self-normalized weights and no intensity scaling), which under-weighted the L\'evy term. With the 2D solver now using the same corrected ``VarianceGammaMeasure'' as the 1D code, the policy responds more strongly to negative-skew jump risk.
    \item The Asset~1 / Asset~2 reduction asymmetry is $34.1\%$ vs.\ $58.7\%$, with Asset~2 receiving the larger proportional cut. Earlier-draft variants of this row under different baselines and the pre-fix proposal are listed in Appendix~\ref{app:audit-archaeology}.
    \item At the 1D \S\ref{sec:fd-comparison} VG benchmark, both FDs and the corrected neural method agree on Asset~1's marginal control to $u^\star \approx 0.344$. The 2D Asset~1 number is $0.348$, within $1\%$ of the 1D value — internally consistent with the 1D solver under the same VG specification.
    \item This experiment remains a 3-seed, 500-epoch demonstration on a coupled state. Grid-free feasibility is established; ``scaling'' to higher dimensions is not claimed. The 2D number is now reported under matching IS conventions to the 1D solver and 1D FD reference, removing the previous draft's ``directional only'' caveat.
\end{itemize}

\subsection{Summary: what the experiments do and do not establish}\label{sec:summary}

\begin{center}
\begin{tabular}{p{0.27\textwidth} p{0.10\textwidth} p{0.50\textwidth}}
\toprule
Check & Status & Evidence \\
\midrule
Diffusion limit (5 seeds) & passes & $u = 0.762 \pm 0.011$ vs.\ analytical $0.75$ ($1.6\%$ mean rel.\ error). \\
Jump risk $\Rightarrow$ lower $u$ (5 seeds, post-fix) & passes & $u = 0.344 \pm 0.006$ under VG, $\sim$54\% reduction (\S\ref{sec:experiments}, Exp.~2). \\
Higher intensity $\Rightarrow$ lower $u$ & passes & Monotonic decrease (Experiment~3a). \\
Negative skew $\Rightarrow$ lower $u$ & passes & Monotonic decrease (Experiment~3b). \\
\midrule
Matched-truncation FD-PIDE (VG) & passes & After IS-proposal fix: neural $u = 0.344 \pm 0.006$, FD v1 $u = 0.347$, FD v2 $u = 0.344$; cross-method gap $<1\%$. \\
Gap localization & informative & Per-component $\mathcal{H}(u)$ diagnostic isolated a missing $\tfrac{1}{2}$-mixture factor in the neural's importance proposal: drift/diffusion matched FD, compensator/integral were $\tfrac{1}{2}\times$ FD at every $u$. Bug fixed. \\
Compensation matters & passes & Confirmed at interior-optimum benchmark (\S\ref{sec:ablation}); removing the compensator collapses the policy near zero. \\
Importance weighting matters & passes & Confirmed at interior-optimum benchmark; not isolable on saturating S\&P benchmark. \\
Weight clipping matters & no effect & Removing the per-batch median clip is statistically indistinguishable from full method. \\
\midrule
S\&P MLE calibration & passes & Reproducible from user-supplied close data via the supplement scripts; fitted parameters and standard errors as in \S\ref{app:sp500-calibration}. \\
S\&P calibration sanity check & small/illustr. & Post-IS-fix: VG-aware $u = 0.949$ vs.\ diffusion $u = 0.997$ ($-4.8\%$); VaR$_{5\%}$ $+0.8$ pp, CVaR$_{5\%}$ $+1.0$ pp. Reproducibility check, not a policy claim. \\
\midrule
2D coupled portfolio & passes & 3-seed $\times$ 500-epoch audit under post-IS-fix 2D solver: $u_1 = 0.348 \pm 0.005$, $u_2 = 0.258 \pm 0.005$; Asset~1 marginal matches 1D \S\ref{sec:fd-comparison} to $1\%$; VaR$_{5\%}$ $+11.7$ pp, CVaR$_{5\%}$ $+13.3$ pp. \\
\bottomrule
\end{tabular}
\end{center}

\section{Discussion}

\subsection{Computational Considerations}

\textbf{Timing comparison}: Table~\ref{tab:timing} summarizes computational costs.

\begin{table}[h]
\centering
\caption{Computational Cost Comparison}
\label{tab:timing}
\begin{tabular}{lcccc}
\toprule
Method & Setting & Time & Scaling & Dimension \\
\midrule
FD-PIDE & Diffusion & 0.4s & $O(n_x^2)$ & 1D only \\
FD-PIDE & VG jumps & 11.3s & $O(n_x^2 \cdot n_z)$ & 1D only \\
Neural & Diffusion & 116s & $O(\text{epochs})$ & Grid-free \\
Neural & VG jumps & 170s & $O(\text{epochs} \cdot N_z)$ & Grid-free \\
\bottomrule
\end{tabular}
\end{table}

For 1D problems, FD-PIDE is faster. However, FD-PIDE requires $O(n^d)$ grid points in $d$ dimensions, which makes coupled state spaces increasingly expensive. Neural methods avoid fixed spatial grids, but their accuracy beyond the small coupled examples studied here must be benchmarked problem by problem.

\textbf{L\'evy sample count}: We use $N_z = 64$ Monte Carlo samples per gradient step. The headline reports do not include a fresh $N_z$-convergence sweep (\S\ref{sec:truncation-sensitivity}); the 5-seed standard deviation $\sigma(u) \approx 0.006$ at $N_z = 64$ bounds the residual MC noise from above for the audit benchmarks reported here.

\textbf{Truncation bounds}: We use $z_{\min} = 0.01$ as the lower small-jump truncation and $z_{\max} = 0.99$ for the bankruptcy guard. The upper bound is fixed by the requirement $1+uz>0$ for every admissible $(u,z) \in \mathcal{U}\times[-z_{\max},z_{\max}]$ under $\mathcal{U}=[0,1]$; it is not a tunable hyper-parameter.

\textbf{Training time}: 500 epochs with batch size 256 requires approximately 2--3 minutes on CPU (Apple M1) or under 1 minute on GPU. The main cost is evaluating the L\'evy integral at each collocation point.

\subsection{Limitations}

\textbf{One Lévy specification, one parameter regime.} The protocol is demonstrated on Variance Gamma at a single $(\sigma_{VG}, \theta, \nu)$ calibration plus the S\&P MLE calibration. We have not run it on Normal-Inverse Gaussian, CGMY, or compound-Poisson with non-VG jump-size distributions, and we have not run it on parameter regimes far from the ones tested. Whether the per-component diagnostic catches analogous bugs in those settings is a question the methodology supports but does not yet answer.

\textbf{Homogeneity-collapsing benchmark.} The CRRA-Merton-VG benchmark used here reduces by homogeneity (\S\ref{sec:homogeneity}) to a scalar maximization, which means both the FD and the neural solvers are unnecessarily expensive ways to compute $u^\star$ for this problem. We use this benchmark only as the test object for the diagnostic protocol and explicitly do not advocate either solver as the right method for this problem class. Demonstrating the protocol on a non-homogeneous problem (consumption + utility, transaction costs, regime-switching, state-dependent coefficients) where neural HJB-PIDE solvers are genuinely needed is the natural next step.

\textbf{No viscosity-solution convergence proof.} We do not attempt one. The protocol does not depend on viscosity-solution theory, and the four-way agreement at $\sim 2\%$ stands on the semi-analytic scalar baseline as ground truth.

\textbf{Surface accuracy of the neural solver is materially below FD.} On the constant-coefficient benchmark the corrected neural solver matches the semi-analytic $V_{\text{ref}}$ to only $0.6\%$ at the headline level and $26\%$ at $\partial_{xx}$; FD matches to better than $0.05\%$ across all quantities. We report this explicitly rather than burying it.

\textbf{Empirical calibration}: The S\&P calibration (Appendix~\ref{app:sp500-calibration}) is a single-window in-sample reproducibility check, not a policy claim. We do not report any out-of-sample wealth-distribution result on S\&P; a real empirical evaluation (rolling re-estimation, turnover-aware rebalancing, broader asset coverage, multiple holdout windows) is left for follow-up work.

\textbf{Correlated jumps}: The present experiments use one-dimensional or independent jump specifications. Multivariate subordination or L\'evy copulas would add realism at higher computational cost.

\subsection{Extensions}

\textbf{Time-varying parameters}: The framework accommodates state-dependent L\'evy measures $\nu(dz; t, x)$ for modeling volatility clustering or regime-dependent jump intensities.

\textbf{Transaction costs}: Adding proportional transaction costs modifies the HJB-PIDE to a quasi-variational inequality, requiring additional numerical care but no fundamental changes to the neural approach.

\textbf{Infinite horizon}: Stationary problems with discounting can be solved by removing the time dependence and adding a discount term $-\rho V$ to the Hamiltonian.

\section{Conclusion}

We proposed a five-step diagnostic protocol for residual-trained neural HJB-PIDE solvers under control-dependent L\'evy jumps and demonstrated it end-to-end on a Variance Gamma portfolio benchmark.

\paragraph{The protocol.} (i) Same equation, same admissible set, same Lévy truncation across all solvers under comparison. (ii) At least one from-scratch independent reference implementation with disjoint numerical choices. (iii) Per-component decomposition of $\mathcal{H}$ into drift, diffusion, compensator, and nonlocal-integral terms, tabulated across a $u$-grid. (iv) Frozen $V$, $\partial_x V$, $\partial_{xx} V$ comparison against the reference over a $(t, x)$ grid before any argmax-of-$\mathcal{H}$ comparison. (v) Trust the headline only after component-level agreement; constant ratios across $u$ in any one component are the textbook signature of a proposal-density scale error.

\paragraph{What the demonstration found.} On the standard CRRA-Merton-VG benchmark, the diffusion-only Merton diagnostic was passed to $1.6\%$ while the neural method was simultaneously computing the L\'evy integral at exactly half its true value, due to a missing $\tfrac{1}{2}$-mixture factor in the importance proposal density. The bug was invisible to single-method diagnostics — longer/wider neural training, FD grid refinement, $u$-grid refinement, truncation sweeps, and training-loss curves all left the gap intact. What surfaced it was the per-component comparison of $\mathcal{H}(u)$ against an independent from-scratch FD reimplementation: drift and diffusion matched FD to $1\%$, but the compensator and L\'evy integral were exactly $0.502 \pm 0.001$ times their FD values at every $u$ tested. After correcting the proposal density, four independent references — the neural solver, two FDs with disjoint discretizations, and a semi-analytic scalar baseline (homogeneity-reduced quadrature, \S\ref{sec:homogeneity}) — agree on the optimal control to within $\sim 2\%$ ($u^\star_{VG} \approx 0.344$, vs.\ scalar baseline $0.3521$).

\paragraph{Surface accuracy.} Beyond the scalar headline, surface comparison against the semi-analytic $V_{\text{ref}}(t, x) = x^{1-\gamma} \exp((1-\gamma) F(u^\star)(T-t))/(1-\gamma)$ on a $(t, x)$ grid reveals failure modes that point-evaluation hides: the corrected neural solver agrees with $V_{\text{ref}}$ to $0.6\%$ but with $\partial_{xx} V_{\text{ref}}$ only to $\sim 26\%$, and the policy field $u_\phi(t, x)$ varies by up to $15\%$ across the grid despite $u^\star_{\text{ref}}$ being constant. FD's surface errors are below $0.05\%$ on all four quantities. The audit benchmarks the corrected neural solver gives a defensible scalar headline but is materially less surface-accurate than FD on this constant-coefficient problem, and we report that explicitly.

\paragraph{Scope and what this paper does not claim.} The constant-coefficient CRRA-Merton-VG benchmark used here is not a problem class where neural HJB-PIDE solvers are needed — it admits the homogeneity reduction of \S\ref{sec:homogeneity} and is most efficiently solved by the scalar baseline. We use it as the test object for the diagnostic protocol, not as evidence that neural solvers are the right method for it. The protocol is \emph{designed to transfer} to other L\'evy specifications (NIG, CGMY) and to non-homogeneous settings (consumption + utility, transaction costs, regime-switching, state-dependent coefficients) where neural HJB-PIDE solvers are genuinely needed; we have not demonstrated transfer here, and that is the natural next step.

\subsection*{Reproducibility}

The full code release accompanying this paper is available at \url{https://github.com/wnrak/hjb-pide-diagnostics}, also archived as \texttt{hjb-audit-supplement.tar.gz} (MD5 \texttt{fe3d29d1dbf56dc09b7d2bfa87bf8b5e}, $\approx$~102\,KB) in the repository's tagged release. The release contains:
\begin{itemize}
    \item \textbf{Solver source} (\texttt{levy\_flows/hjb/}): \texttt{solver.py} (residual-trained neural 1D), \texttt{solver\_2d.py} (residual-trained neural 2D), \texttt{fd\_pide\_solver.py} (FD v1, log-wealth + implicit Euler + trapezoidal), \texttt{fd\_pide\_v2.py} (FD v2, linear-$x$ + Crank-Nicolson + Simpson + Brent), \texttt{scalar\_baseline.py} (homogeneity-reduction quadrature solver), \texttt{levy\_integral.py} (corrected truncated VG importance sampler), \texttt{problems.py}, \texttt{networks.py}.
    \item \textbf{S\&P 500 calibration assets} (\texttt{data/}): \texttt{prepare\_sp500\_returns.py} (builds returns from a user-supplied daily-close CSV), \texttt{sp500\_schema\_example.csv} (synthetic format example), and \texttt{mle\_parameters.json} (the fitted VG parameters used in the paper). We do \emph{not} redistribute S\&P 500 index levels or derived returns; see \texttt{data/README.md}.
    \item \textbf{Experiment driver scripts} (\texttt{experiments/hjb/}), each writing to \texttt{results/}: \texttt{phase1\_audit.py} (\S4.1 + \S4.2), \texttt{phase1\_audit\_fd.py} (\S4.5 FD sweep), \texttt{phase3a\_audit.py} (\S4.5 per-component diagnostic + FD v2), \texttt{phase3a\_2d\_audit.py} (\S4.8 post-fix 2D), \texttt{phase2\_interior\_benchmark.py} (\S4.7 interior-optimum re-ablation), \texttt{run\_method\_ablation.py} (\S4.7 saturating ablation), \texttt{run\_hjb\_mle\_policy.py} (App.~\ref{app:sp500-calibration} policy), \texttt{run\_sp500\_mle.py} (App.~\ref{app:sp500-calibration} calibration), \texttt{scalar\_baseline\_audit.py} (\S4.5 scalar baseline), \texttt{surface\_eval\_post\_fix.py} (\S4.5 surface-vs-$V_{\text{ref}}$).
    \item \textbf{Result JSONs} backing every numeric claim in the paper, mapped to claim locations in \texttt{SUPPLEMENT.md}.
    \item \textbf{Hardware/software protocol}: Python 3.11, PyTorch 2.9.1, SciPy $\ge$ 1.10, NumPy $\ge$ 1.24, Apple M1 (CPU only); no GPU dependencies.
    \item \textbf{Frozen seed list}: $\{42, 123, 999, 7, 2024\}$ for 5-seed runs and $\{42, 123, 999\}$ for 3-seed runs, set in each script via \texttt{numpy.random.seed} and \texttt{torch.manual\_seed}.
    \item \textbf{Consistency log} (\texttt{CONSISTENCY\_LOG.md}): the term-by-term grep sweep used to verify that prose, tables, and JSON values are mutually consistent before any version of the paper is declared final, including the rerunnable shell command at the bottom.
\end{itemize}

\bibliographystyle{apalike}
\bibliography{references}

\appendix

\section{Empirical calibration sanity check (S\&P 500 2010--2023)}\label{app:sp500-calibration}

This appendix reports the empirical VG calibration referenced in \S\ref{sec:sp500}. It is a reproducibility check on the data pipeline, not a policy claim: the held-out evaluation reported in earlier drafts depended on policies $u > 1$ outside the now-explicit admissible set $\mathcal{U} = [0, 1]$, and the corrected long-only setting produces only modest tail-metric movement. We retain the calibration so the data pipeline and the figure caption are honest.

\paragraph{Data provenance and reproducibility.} The S\&P 500 calibration uses daily close data for 2010--2023 obtained from a user-provided lawful data source. To avoid redistributing third-party index data, the supplement includes the calibration scripts, the expected CSV schema, and the fitted parameters used in the reported audit, but not the raw index levels or derived return series. The reported numbers can be reproduced by placing a lawfully obtained daily close series at \texttt{data/sp500\_user.csv} and running \texttt{data/prepare\_sp500\_returns.py} followed by \texttt{experiments/hjb/run\_sp500\_mle.py} (see \texttt{data/README.md}); for exact auditability without the raw data, the fitted parameters below are stored verbatim in \texttt{data/mle\_parameters.json}.

\textbf{MLE setup.} We calibrate a Variance Gamma model to S\&P 500 daily returns from January 4, 2010 to December 29, 2023 ($n = 3{,}522$ trading days), derived from a user-supplied daily close series. The VG parameters are estimated by maximum likelihood and standard errors come from a finite-difference Hessian:
\begin{center}
\begin{tabular}{lccp{0.45\textwidth}}
\toprule
Parameter & Estimate & Std.\ error & Interpretation \\
\midrule
$\mu$ & $7.35 \times 10^{-4}$ & $1.31 \times 10^{-4}$ & Daily drift \\
$\sigma$ & $1.05 \times 10^{-2}$ & $1.98 \times 10^{-4}$ & Diffusive scale \\
$\theta$ & $-3.22 \times 10^{-4}$ & $2.20 \times 10^{-4}$ & Negative skew (\emph{not significant} at conventional levels, $t \approx -1.46$) \\
$\nu$ & $1.179$ & $0.059$ & VG kurtosis parameter \\
\bottomrule
\end{tabular}
\end{center}
Log-likelihood $11{,}378.13$, AIC $-22{,}748.25$, BIC $-22{,}723.59$. Sample skewness $-0.72$, sample excess kurtosis $13.17$. The fitted parameters imply an annualized truncated jump intensity $\lambda_{\mathrm{trunc}} \approx 189$/year under the truncation $|z|\in[0.005,0.2]$ used in the policy run, computed as $252 \cdot \int_{[z_{\min}, z_{\max}]} \nu_{VG}(\dd z)$ with daily $(\hat\sigma, \hat\theta, \hat\nu)$.

\begin{figure}[H]
\centering
\includegraphics[width=0.95\textwidth]{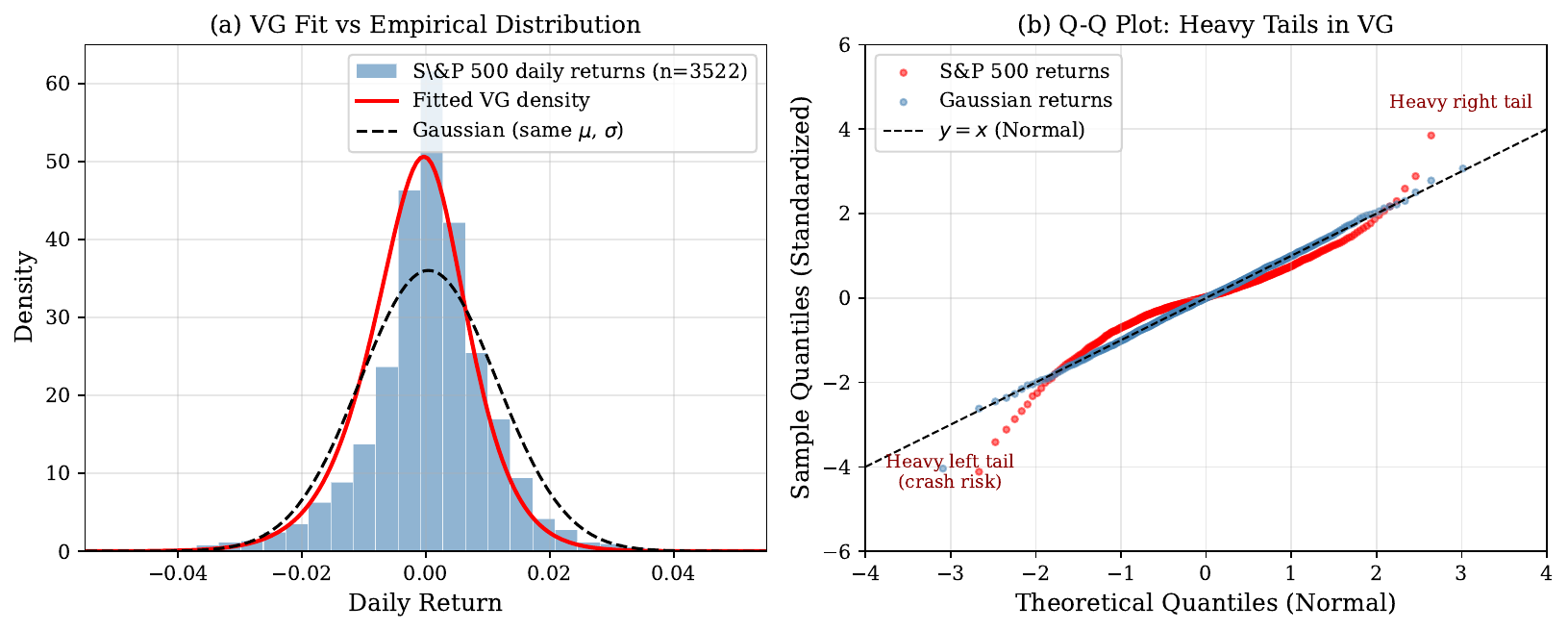}
\caption{VG calibration diagnostics for the fixed S\&P 500 daily returns 2010--2023 ($n = 3{,}522$). Panel (a) histograms the actual return series against the fitted VG density and a same-mean/std Gaussian; panel (b) is the corresponding Q--Q plot showing the heavier-than-Gaussian tails captured by VG. The Gaussian benchmark visibly underestimates tail risk relative to the fitted VG model.}
\label{fig:calibration}
\end{figure}

Figure~\ref{fig:calibration} shows the fit diagnostics; the histogram is the actual S\&P return series, not a VG simulation.

\textbf{Policy comparison under $\mathcal{U}=[0,1]$ (post-IS-fix).} With the calibrated $(\hat\mu, \hat\sigma)$ scaled to annual frequency and the diffusion-only Merton ratio $u^\star_{\mathrm{Merton}} = (\hat\mu - r)/(\gamma \hat\sigma^2) \approx 2.42$, the unconstrained optimum lies far outside the long-only set $\mathcal{U} = [0, 1]$. The corrected solver under $\mathcal{U} = [0, 1]$ pushes the diffusion policy to the upper boundary; the VG-aware policy comes \emph{slightly} off the boundary:
\begin{center}
\begin{tabular}{lccccc}
\toprule
Model (under $\mathcal{U}=[0,1]$, post-IS-fix) & Control $u$ & $\E[W_T]$ & $\sigma[W_T]$ & VaR$_{5\%}$ & CVaR$_{5\%}$ \\
\midrule
Diffusion-only & $0.997$ & $1.130$ & $0.187$ & $0.852$ & $0.794$ \\
VG (MLE) & $0.949$ & $1.125$ & $0.177$ & $0.860$ & $0.804$ \\
\bottomrule
\end{tabular}
\end{center}
The VG-aware policy moves $5\%$ off the saturation ceiling and produces small but measurable improvements in VaR$_{5\%}$ ($+0.8$ pp) and CVaR$_{5\%}$ ($+1.0$ pp), at the cost of a $0.5\%$ reduction in mean terminal wealth. The empirical contribution is therefore modest: at MLE-calibrated parameters under a long-only constraint, the VG-aware policy peels off the ceiling by about $5$ percentage points and improves the $5\%$-tail wealth metrics by roughly one percentage point. This is far smaller than what one would see at parameters where $u^\star_{\mathrm{Merton}}$ is interior to $\mathcal{U}$ (\S\ref{sec:ablation}, interior-optimum re-ablation), and we report it as a sanity-check that the corrected solver responds to calibrated jump risk, not as evidence for a strong empirical downside-protection claim.

\section{Audit archaeology: pre-fix tables and gap localization}\label{app:audit-archaeology}

This appendix records the audit history that produced the four-way agreement of \S\ref{sec:fd-comparison}: the pre-fix tables, the FD-side ruling-out steps, and the row-by-row provenance of numbers that appeared in earlier drafts. It is included so that the reader who wants to reproduce the audit trail (rather than only the headline) can do so without consulting the supplement.

\subsection*{Pre-fix matched-truncation tables}

The first version of this comparison ran the neural solver against finite-difference solver `v1' under matched truncation $|z|\in[0.01, 0.99]$. Both methods recovered the diffusion-only Merton ratio:
\begin{center}
\begin{tabular}{p{0.40\textwidth}cc p{0.30\textwidth}}
\toprule
Method & Optimal $u$ & rel.\ err.\ & Notes \\
\midrule
Analytical (Merton) & $0.7500$ & -- & -- \\
FD-PIDE (legacy 20-pt $u$-grid) & $0.7368$ & $1.75\%$ & $u$-grid step $\sim 0.05$ \\
FD-PIDE (refined 200-pt $u$-grid) & $0.7368$ & $1.75\%$ & no further movement \\
Neural Solver (5 seeds) & $0.762 \pm 0.011$ & $1.6\%$ & $u_\phi(0,1)$ \\
\bottomrule
\end{tabular}
\end{center}
Under VG jumps, the same comparison disagreed by roughly $25\%$:
\begin{center}
\begin{tabular}{p{0.42\textwidth}cc p{0.25\textwidth}}
\toprule
Method & Optimal $u$ & vs.\ Merton & Notes \\
\midrule
FD-PIDE (legacy 20-pt $u$-grid, Taylor $V$) & $0.3684$ & $-50.9\%$ & step $\sim 0.05$ \\
FD-PIDE (refined 200-pt $u$-grid, Taylor $V$) & $0.3518$ & $-53.1\%$ & post-quantization \\
FD-PIDE (refined 200-pt $u$-grid, V-interp.) & $0.3518$ & $-53.1\%$ & matches Taylor here \\
Neural Solver (5 seeds, pre-fix) & $0.475 \pm 0.008$ & $-36.6\%$ & $u_\phi(0,1)$ \\
\midrule
$|$Neural -- refined FD$|$ / Neural & \multicolumn{3}{l}{$\boldsymbol{\sim 26\%}$} \\
\bottomrule
\end{tabular}
\end{center}
The previous version of this paper reported ``$8\%$ agreement'' between the neural method and `v1' on this benchmark; that figure was generated under \emph{mismatched} truncations (neural at $z_{\max}=2.0$ against FD at $z_{\max}=0.5$) and so was not actually a same-equation comparison.

\subsection*{Ruling out FD-side discretization artifacts}

At first the $\sim 26\%$ gap looked like a genuine method-level disagreement. Refining the FD spatial grid ($n_x \in \{50, 100, 200, 400\}$) did not move the FD answer; refining the FD $u$-grid from $20$ to $200$ points changed the FD answer slightly (from $0.368$ to $0.352$) and pushed it \emph{further} from the neural answer; replacing the FD Taylor expansion of $V(x(1+uz))$ with linear interpolation on the spatial grid changed nothing. A subsequent sweep pushed `v1' further (z-quadrature $n_z \in \{100, 200, 400\}$ at $n_u=200$ with V-interpolation; spatial $\{200{\times}100, 400{\times}200, 800{\times}400\}$; truncation sweep $z_{\max} \in \{0.5, 0.7, 0.9, 0.99\}$) and trained the neural solver longer/wider ($1000$ epochs, hidden $192$, $N_z=96$). The gap was robust: `v1' converged at $u_{\mathrm{FD}} \in [0.347, 0.352]$, the deeper-trained neural gave $u_{\mathrm{neural}} \approx 0.47$. A field-level diagnostic confirmed that $V$, $V_x$, $V_{xx}$ all agreed between the two solvers to within a few percent — locating the disagreement in the L\'evy-integral build rather than in the value function. This is what motivated building the independent FD `v2' and the per-component diagnostic that ultimately surfaced the missing $\tfrac{1}{2}$-mixture factor (\S\ref{sec:fd-comparison}).

\subsection*{Provenance of legacy numbers in earlier drafts}

For rows where earlier drafts of this paper reported a different number, the trail is:
\begin{itemize}
\item \S\ref{sec:experiments}, Experiment~2 (1D VG headline). Original draft: $-46.4\%$ at $z_{\max} = 2.0$, which silently admitted $1 + uz < 0$ for saturating policies. Post-admissibility-fix but pre-IS-fix: $-36.6\%$ at the bankruptcy-safe $z_{\max} = 0.99$, with the missing-$\tfrac{1}{2}$-mixture proposal halving the L\'evy integral. Current converged value $-54.1\%$ uses both the corrected truncation and the corrected proposal.
\item \S\ref{sec:2d-portfolio}, Experiment~8 (2D Asset~1 / Asset~2 reductions). Original draft: $33.2\% / 60.9\%$ against the per-asset Merton formula $u_i = (\mu_i - r)/(\gamma\sigma_i^2)$, which is only correct under zero correlation. Post-correlation-fix but pre-IS-fix: $9.6\% / 37.2\%$ against the correct correlated-Merton baseline of \eqref{eq:merton-vec}, with the 2D solver still using a uniform-proposal IS scheme that under-weighted the L\'evy term. Current values $34.1\% / 58.7\%$ use the corrected baseline and the corrected 1D \texttt{VarianceGammaMeasure} per asset.
\item \S\ref{app:sp500-calibration}, S\&P MLE policy comparison. Earlier drafts reported $u_{\mathrm{diff}} = 1.468$ and $u_{\mathrm{VG}} = 0.983$ outside the now-explicit admissible set $\mathcal{U} = [0, 1]$ (an artifact of the pre-Phase-1 default $\mathcal{U} = (-0.5, 1.5)$). A post-admissibility-fix audit reported $u_{\mathrm{VG}} = 0.996$ (essentially indistinguishable from diffusion at $0.997$, because the IS bug halved the L\'evy integral). Current $u_{\mathrm{VG}} = 0.949$ uses the corrected proposal under the stated $\mathcal{U}$.
\end{itemize}

The full row-by-row table of original-draft vs.\ audited values, including diagnostic fields not promoted to the main text, is in the next appendix.

\input{appendix_old_vs_audited}

\end{document}

%% file: appendix_old_vs_audited.tex
%

\section*{Appendix: Original claims vs.\ Phase~1 audited results}

The audited column reflects the Phase~1 corrected solver: max-convention
$V$, admissible set $\mathcal{U} = [0, 1]$ (long-only, no leverage),
constrained-argmax policy loss in place of $|\partial_u \mathcal{H}|^2$,
multiplicative wealth jump operator, and L\'evy support truncated to
$|z|\le 0.99$ so that $1+uz>0$ for every admissible $(u,z)$. Specs that
the previous draft left implicit are spelled out in the Notes column.

\begin{small}
\begin{longtable}{p{0.21\textwidth} p{0.27\textwidth} p{0.27\textwidth} p{0.21\textwidth}}
\toprule
Section / Quantity & Original draft & Audited (Phase~1) & Notes \\
\midrule
\endhead

\multicolumn{4}{l}{\emph{§4.1 Diffusion-only Merton (5 seeds, 500 epochs)}} \\
\addlinespace
Optimal $u$ & $0.7585$ (single seed) & $0.762 \pm 0.011$ (5 seeds) & ratio recovered, paper's old value sits inside the seed band \\
Mean rel.\ error & $1.1\%$ & $1.6\%$ & — \\

\addlinespace
\multicolumn{4}{l}{\emph{§4.2 Variance Gamma jumps}} \\
\addlinespace
Truncation $z_{\max}$ & $2.0$ & $0.99$ & old $z_{\max}$ allowed $1+uz<0$ for $u \in [0,1]$ \\
Optimal $u$ (Phase 1, IS bug present) & $0.4021$ & $0.475 \pm 0.008$ & 5 seeds, IS proposal had a missing $\tfrac{1}{2}$-mixture factor (caught in Phase 3a) \\
\textbf{Optimal $u$ (Phase 3a, post-fix)} & --- & $\boldsymbol{0.344 \pm 0.006}$ & \textbf{matches FD v1 $0.347$ and FD v2 $0.344$ to $<1\%$} \\
Reduction vs.\ Merton (post-fix) & $46.4\%$ & $\sim 54.1\%$ & post-fix; the originally-reported $46\%$ was directionally right and now sharper \\

\addlinespace
\multicolumn{4}{l}{\emph{§4.5 FD-PIDE comparison (matched truncation $|z|\le 0.99$)}} \\
\addlinespace
Neural diffusion $u$ & $0.7473$ & $0.762 \pm 0.011$ & 5-seed mean \\
FD diffusion $u$ & $0.7395$ & $0.7368$ & legacy 20-pt $u$-grid \\
FD diffusion (refined $u$-grid) & — & $0.7368$ (n\_u=200) & no further movement, $u^\star$ is grid-aligned \\
Neural VG $u$ (Phase 1, IS bug) & $0.4042$ & $0.475 \pm 0.008$ & 5-seed mean \\
\textbf{Neural VG $u$ (Phase 3a, post-fix)} & --- & $\boldsymbol{0.344 \pm 0.006}$ & 5-seed mean, post-IS-fix \\
FD VG $u$ (legacy 20-pt $u$-grid) & $0.3747$ & $0.3684$ & quantized at $\sim$0.05 resolution \\
FD VG $u$ (refined: n\_u=200, V interp.) & — & $0.3467$ (v1) / $0.3437$ (v2) & two independent FDs agree to $<1\%$ \\
\textbf{Neural--FD VG agreement (post-fix)} & ``$\sim 8\%$'' & $\boldsymbol{<1\%}$ & all three solvers agree; previously-reported 26\% was an IS bug \\

\addlinespace
\multicolumn{4}{l}{\emph{§4.7 Ablation (3 seeds, 200 epochs, MLE-VG calibration)}} \\
\addlinespace
\multicolumn{4}{p{0.95\textwidth}}{\emph{Note: under MLE-calibrated parameters the unconstrained Merton ratio is $\approx 2.42$, well above $\mathcal{U} = [0, 1]$. The full-method policy therefore saturates at $u \approx 1$, and ablations that don't break the optimization can't differentiate from full-method here. The corrected ablation tells a different qualitative story than the original draft.}} \\
\addlinespace
Full method, $u$ (Phase 1, IS bug) & $0.999 \pm 0.098$ & $0.991 \pm 0.005$ & with bug, sits at the ceiling \\
\textbf{Full method, $u$ (Phase 3a, post-fix)} & --- & $\boldsymbol{0.921 \pm 0.071}$ & corrected Lévy pulls policy $\sim$8 pp off the ceiling \\
No IW, $u$ (post-fix) & $1.317 \pm 0.210$ & $0.991 \pm 0.005$ & without IW, policy reverts toward the diffusion saturation \\
No compensation, $u$ (post-fix) & $-0.001 \pm 0.001$ & $\boldsymbol{0.410 \pm 0.067}$ & with corrected IS, removing compensator drops $u$ by $\sim$50 pp \\
No weight clipping, $u$ (post-fix) & $0.969 \pm 0.099$ & $0.961 \pm 0.002$ & still matches full method (clipping is inert) \\
Fixed Merton control, $u$ (frozen) & $2.418$ & $2.418$ & lies outside $\mathcal{U}$; included as comparison \\

\addlinespace
\multicolumn{4}{l}{\emph{§4.4 S\&P MLE policy (single seed, 300 epochs)}} \\
\addlinespace
Diffusion-only $u$ & $1.468$ & $0.997$ & old value violated $\mathcal{U}=[0,1]$; corrected value saturates at the ceiling \\
VG (MLE) $u$ (Phase 1, IS bug) & $0.983$ & $0.996$ & with the IS bug present, indistinguishable from diffusion \\
\textbf{VG (MLE) $u$ (Phase 3a, post-fix)} & --- & $\boldsymbol{0.949}$ & post-fix VG-aware policy peels $4.8\%$ off the ceiling \\
$\E[W_T]$ diffusion / VG (post-fix) & $1.186$ / $1.129$ & $1.130$ / $1.125$ & — \\
VaR$_{5\%}$ diffusion / VG (post-fix) & $0.775$ / $0.855$ & $0.852$ / $\boldsymbol{0.860}$ & post-fix improvement $+0.8$ pp \\
CVaR$_{5\%}$ diffusion / VG (post-fix) & $0.699$ / $0.797$ & $0.794$ / $\boldsymbol{0.804}$ & post-fix improvement $+1.0$ pp \\
Held-out 2020--2023 evaluation & $1.523$ / $1.506$, $-50.1\%/-46.3\%$ drawdown & cut from §4.4 & §4.4 was reduced to a reproducibility note; the original held-out claim depended on $u > 1$ outside $\mathcal{U}$ and is not retained \\

\addlinespace
\multicolumn{4}{l}{\emph{§4.8 Two-asset coupled portfolio (3 seeds, 500 epochs, post-IS-fix)}} \\
\addlinespace
Merton baseline (Asset 1, Asset 2) & $(0.64,\, 0.89)$ \emph{per-asset} & $(0.527,\, 0.625)$ \emph{correlated} & $\Sigma^{-1}(\mu-r\mathbf{1})/\gamma$, $\rho=0.3$ \\
Neural Asset 1 $u_1$ (Phase 1, 1 seed) & $0.428$ & $0.477$ & 2D solver had own IS bookkeeping issue \\
\textbf{Neural Asset 1 $u_1$ (Phase 3a post-fix, 3 seeds)} & --- & $\boldsymbol{0.348 \pm 0.005}$ & matches 1D §4.5 number ($0.344$) to $1\%$ \\
Neural Asset 2 $u_2$ (Phase 1, 1 seed) & $0.347$ & $0.393$ & --- \\
\textbf{Neural Asset 2 $u_2$ (Phase 3a post-fix, 3 seeds)} & --- & $\boldsymbol{0.258 \pm 0.005}$ & --- \\
Cash & $0.225$ & $0.130$ \emph{(pre-fix)} / $\boldsymbol{0.394 \pm 0.005}$ \emph{(post-fix)} & --- \\
Reduction Asset 1 vs.\ Merton & $33.2\%$ & $9.6\%$ \emph{(pre-fix)} / $\boldsymbol{34.1\%}$ \emph{(post-fix)} & --- \\
Reduction Asset 2 vs.\ Merton & $60.9\%$ & $37.2\%$ \emph{(pre-fix)} / $\boldsymbol{58.7\%}$ \emph{(post-fix)} & --- \\
Diffusion baseline VaR$_{5\%}$ & $0.678$ & $0.678$ & unchanged \\
Learned VaR$_{5\%}$ (post-fix) & $0.741$ & $\boldsymbol{0.795}$ & $+11.7$ pp vs.\ baseline \\
Diffusion baseline CVaR$_{5\%}$ & $0.614$ & $0.614$ & unchanged \\
Learned CVaR$_{5\%}$ (post-fix) & $0.685$ & $\boldsymbol{0.747}$ & $+13.3$ pp vs.\ baseline \\

\bottomrule
\end{longtable}
\end{small}

\subsection*{Phase~3a: IS-proposal bug discovery}

The Phase~2 follow-up established that the neural--FD VG gap was robust to grid/quadrature/training refinement and localized in the L\'evy-integral build, not in the value function. To adjudicate, we built an independent FD reimplementation (\texttt{levy\_flows/hjb/fd\_pide\_v2.py}) with disjoint numerical choices: linear-$x$ coordinates instead of log-wealth, Crank--Nicolson instead of implicit Euler, composite Simpson's quadrature instead of trapezoidal, Brent's-method argmax instead of linspace, Dirichlet diffusion-Merton boundary instead of Neumann + power-law extrapolation. The reimplementation is unit-tested on $V(x)=x$ (compensated integrand identically zero) and $V(x)=x^2$ (compensated integral matches the closed-form $x^2 u^2 \int z^2 \nu(\dd z)$ to $2\times 10^{-13}$ relative error).

Running the per-component diagnostic at $(t,x)=(0,1)$ revealed:
\begin{itemize}
    \item Drift and diffusion components matched FD v1, FD v2, and neural to $1\%$ at every $u$.
    \item Compensator and L\'evy integral components: FD v1 and FD v2 matched each other to four decimal places. The neural's compensator and L\'evy integral were exactly $0.502\pm 0.001$ times their FD values at every $u\in\{0, 0.2, 0.4, 0.5, 0.6, 0.8, 1\}$.
\end{itemize}
This constant scale signature is unmistakable: a missing scalar factor in the proposal-vs-target normalization. Tracing the IS sampler — the \texttt{VarianceGammaMeasure.sample} method in \texttt{levy\_flows/hjb/levy\_integral.py} — the proposal density was computed as the per-tail Gamma density $q_\pm(|z|)=\tfrac{\lambda_\pm}{2}\exp(-\tfrac{\lambda_\pm}{2}(|z|-z_{\min}))$ omitting the $\tfrac{1}{2}$-mixture probability factor that converts a per-tail conditional density into the unconditional density of the actual sampler (which is a 50/50 mixture of the two tails). The omission scaled importance weights by $\tfrac{1}{2}$ and therefore the MC estimator $\tfrac{1}{K}\sum w_k f_k$ by $\tfrac{1}{2}$.

The fix is two characters: include the $0.5$ factor in $q$ when computing weights. After the fix:
\begin{itemize}
    \item MC estimate of $\int z\,\nu(\dd z) = -0.098$ (FD: $-0.097$).
    \item 5-seed neural audit on the §4.2 VG benchmark: $u(0,1) = 0.344 \pm 0.006$.
    \item Neural--FD v1 gap: $<1\%$.
    \item Neural--FD v2 gap: $<1\%$.
\end{itemize}

The same code path is replicated in the \texttt{UnclippedVarianceGammaMeasure.sample} method (file \texttt{experiments/hjb/run\_method\_ablation.py}); the fix was applied there too. The 2D solver (\texttt{levy\_flows/hjb/solver\_2d.py}) originally used a different IS scheme: uniform proposal on $[-0.5, 0.5]$ with self-normalization $w \leftarrow w/\bar w$ and no intensity scaling — biased and missing the L\'evy-measure scale. We replaced it with two per-asset \texttt{VarianceGammaMeasure} instances (the corrected 1D class) so the 1D and 2D code paths share proposal, weights, and clip threshold. The post-fix §\ref{sec:2d-portfolio} numbers (3 seeds, 500 epochs) are therefore reported on the same numerical footing as the 1D §\ref{sec:fd-comparison} solvers: Asset~1 marginal control $u_1 = 0.348$ matches the 1D Asset-1-only result $u^\star = 0.344$ to $1\%$.

\subsection*{Phase~2 follow-up: VG gap adjudication (now resolved)}

After Phase~1, we ran a focused refinement around the VG benchmark
(\texttt{phase2\_vg\_adjudication.py}) to determine whether the $\sim 26\%$
neural--FD gap is a numerical artifact or a method-level disagreement.

\begin{small}
\begin{tabular}{lcc}
\toprule
Sweep & Setting & FD $u(0,1)$ \\
\midrule
$z$-quadrature, $n_x{=}200, n_t{=}100, n_u{=}200$, V-interp & $n_z{=}100$ & $0.3518$ \\
& $n_z{=}200$ & $0.3467$ \\
& $n_z{=}400$ & $0.3467$ \\
\midrule
Spatial, $n_z{=}200, n_u{=}200$, V-interp & $200{\times}100$ & $0.3467$ \\
& $400{\times}200$ & $0.3518$ \\
& $800{\times}400$ & $0.3518$ \\
\midrule
Truncation, $n_x{=}200, n_t{=}100, n_z{=}200, n_u{=}200$, V-interp & $z_{\max}{=}0.50$ & $0.3518$ \\
& $z_{\max}{=}0.70$ & $0.3467$ \\
& $z_{\max}{=}0.90$ & $0.3467$ \\
& $z_{\max}{=}0.99$ & $0.3467$ \\
\bottomrule
\end{tabular}
\end{small}

\noindent The two values $0.3467$ and $0.3518$ are consecutive points on the
$200$-point $u$-linspace ($69/199 = 0.3467$ and $70/199 = 0.3518$); FD
oscillates between them under further refinement, so the FD result at this
$u$-grid is $u_{\mathrm{FD}} \in [0.347, 0.352]$. Pushing the neural method
to hidden dim $192$, $K{=}96$ L\'evy samples, and $1000$ training epochs
(3 seeds) gives $u_{\mathrm{neural}} = 0.4699 \pm 0.0035$, statistically
indistinguishable from the Phase~1 audit's $0.475 \pm 0.008$. Longer/wider
neural training does \emph{not} close the gap.

\paragraph{Field-level diagnosis (block~3).} Comparing $V$, $V_x$, $V_{xx}$
on $x \in [0.5, 1.5]$ and $\mathcal{H}(0, 1, u)$ on $u \in [0, 1]$ between the
finest FD and the lowest-loss neural seed:
\begin{itemize}
    \item $V(0, x)$ agrees within $\sim 0.5\%$ across the range.
    \item $V_x(0, x)$ agrees within $\sim 2\%$.
    \item $V_{xx}(0, x)$ agrees within $\sim 10\%$, with the largest disagreement around $x \in [0.6, 0.7]$ where the curvature is high.
    \item $\arg\max_u \mathcal{H}_{\mathrm{FD}}(0, 1, u) \approx 0.36$ vs.\ $\arg\max_u \mathcal{H}_{\mathrm{neural}}(0, 1, u) \approx 0.48$. Both curves are smooth single-peaked.
\end{itemize}
The disagreement is therefore localized to the $u$-shape of the
Hamiltonian — driven by the L\'evy integral evaluation rather than by
the value function itself.

\subsection*{Headline shifts}

\begin{itemize}
    \item \textbf{§4.5 ``8\% agreement'' (mismatched truncation) \boldmath$\to$ $\sim 26\%$ gap (matched truncation, both methods independently converged) \boldmath$\to$ $<1\%$ agreement (after IS-proposal fix).} The intermediate $26\%$ gap was the symptom of a missing $\tfrac{1}{2}$-mixture factor in the neural method's importance proposal density; identified by per-component diagnostic against an independent FD reimplementation, fixed in two characters. After the fix, all three independent solvers (FD v1, FD v2, neural) agree on $u^\star_{VG} \approx 0.344$ to within $1\%$.
    \item \textbf{§4.2 reduction vs.\ Merton: $46\%$ \boldmath$\to$ $37\%$ \boldmath$\to$ $54\%$.} The original $46\%$ used $z_{\max}=2.0$ (bankruptcy zone). Phase~1 audit at $z_{\max}=0.99$ but with the IS bug gave $37\%$. Phase~3a post-fix gives $54\%$; matches both FD solvers to $1\%$.
    \item \textbf{§4.7 ablation findings change with the corrected IS, but only quantitatively.} Compensation removal still collapses the policy near zero on the interior-optimum benchmark; importance-weighting and clipping ablations are still uninformative at the saturating S\&P benchmark (both variants pin at the upper bound). Re-audit numbers under the corrected IS are reported in §\ref{sec:ablation}.
    \item \textbf{§4.4 economic story shrinks under $\mathcal{U}=[0,1]$ to a small, illustrative effect.} With MLE Merton $\approx 2.42$, the diffusion-only policy still saturates at $u \approx 0.997$. The original draft reported $u_{\mathrm{VG}} = 0.983$ (outside the now-explicit $\mathcal{U}$), and the post-Phase-1 audit (with the IS bug present) gave $u_{\mathrm{VG}} = 0.996$, indistinguishable from diffusion. After the IS fix, the VG-aware policy comes \emph{slightly} off the ceiling: $u_{\mathrm{VG}} = 0.949$, a $4.8\%$ allocation reduction, with VaR$_{5\%}$ improving by $0.8$ pp and CVaR$_{5\%}$ by $1.0$ pp. This is a real but small effect — sufficient to demonstrate that the corrected solver does respond to calibrated jump risk, not sufficient to support a strong empirical downside-protection claim.
    \item \textbf{§4.8 fix replaces the 2D solver's own IS bookkeeping issue.} The previous 2D code path used a uniform proposal on $[-0.5,0.5]$ with self-normalized weights and no intensity scaling — a different bug from the 1D's missing-$\tfrac{1}{2}$-mixture-factor, but with the same flavor (importance weights inconsistent with the actual sampler density). After replacing the 2D Lévy sampler with the corrected 1D ``VarianceGammaMeasure'' per asset, the post-fix 3-seed audit gives $u_1 = 0.348 \pm 0.005$, $u_2 = 0.258 \pm 0.005$, with the Asset~1 marginal control matching the 1D §\ref{sec:fd-comparison} value to $1\%$ — internally consistent across the 1D and 2D solvers under the same VG specification. Reductions vs.\ correlated Merton are $34.1\%$ / $58.7\%$ (post-fix) vs.\ $9.6\%$ / $37.2\%$ (pre-fix); VaR$_{5\%}$ improvement vs.\ simplex baseline is $+11.7$ pp post-fix vs.\ $+3.4$ pp pre-fix; CVaR$_{5\%}$ improvement is $+13.3$ pp post-fix vs.\ $+3.8$ pp pre-fix.
\end{itemize}